\RequirePackage{fix-cm}
\documentclass[twocolumn]{svjour3}          
\smartqed  
\usepackage{graphicx}

\usepackage{times}
\usepackage{graphicx,soul,framed,epsfig,mathptmx,times,
amsmath,amssymb,color,flushend,gensymb,subfigure,
fancyhdr,subfigure,lipsum,epstopdf, amsfonts,cite }
\usepackage[binary-units=true]{siunitx}

\usepackage{setspace}
\usepackage{graphicx,soul,framed,mathptmx,times,amsmath,amssymb,color,
	flushend,gensymb,subfigure,fancyhdr,lipsum, amsfonts}
\usepackage{cite}
\usepackage[binary-units=true]{siunitx}
\usepackage{multirow}
\usepackage{tikz}
\usepackage{url}
\usepackage{adjustbox}
\usepackage{booktabs}
\usetikzlibrary{shapes,arrows}

\usepackage[algoruled,boxed,lined]{algorithm2e}

\DeclareMathOperator{\atantwo}{atan2}        
\DeclareMathAlphabet{\pazocal}{OMS}{zplm}{m}{n}   
\DeclareMathOperator*{\argmin}{arg\,min} 
\graphicspath{{Figures/}}

 \journalname{Autonomous Robots}
\RequirePackage[normalem]{ulem} 
\RequirePackage{color}\definecolor{RED}{rgb}{1,0,0}\definecolor{BLUE}{rgb}{0,0,1} 
\providecommand{\DIFaddbegin}{} 
\providecommand{\DIFaddend}{} 
\providecommand{\DIFdelbegin}{} 
\providecommand{\DIFdelend}{} 
\providecommand{\DIFaddbeginFL}{} 
\providecommand{\DIFaddendFL}{} 
\providecommand{\DIFdelbeginFL}{} 
\providecommand{\DIFdelendFL}{} 
\newcommand{\DIFscaledelfig}{0.5}
\RequirePackage{settobox} 
\RequirePackage{letltxmacro} 
\newsavebox{\DIFdelgraphicsbox} 
\newlength{\DIFdelgraphicswidth} 
\newlength{\DIFdelgraphicsheight} 
\LetLtxMacro{\DIFOincludegraphics}{\includegraphics} 
\newcommand{\DIFaddincludegraphics}[2][]{{\color{blue}\fbox{\DIFOincludegraphics[#1]{#2}}}} 
\newcommand{\DIFdelincludegraphics}[2][]{
\sbox{\DIFdelgraphicsbox}{\DIFOincludegraphics[#1]{#2}}
\settoboxwidth{\DIFdelgraphicswidth}{\DIFdelgraphicsbox} 
\settoboxtotalheight{\DIFdelgraphicsheight}{\DIFdelgraphicsbox} 
\scalebox{\DIFscaledelfig}{
\parbox[b]{\DIFdelgraphicswidth}{\usebox{\DIFdelgraphicsbox}\\[-\baselineskip] \rule{\DIFdelgraphicswidth}{0em}}\llap{\resizebox{\DIFdelgraphicswidth}{\DIFdelgraphicsheight}{
\setlength{\unitlength}{\DIFdelgraphicswidth}
\begin{picture}(1,1)
\thicklines\linethickness{2pt} 
{\color[rgb]{1,0,0}\put(0,0){\framebox(1,1){}}}
{\color[rgb]{1,0,0}\put(0,0){\line( 1,1){1}}}
{\color[rgb]{1,0,0}\put(0,1){\line(1,-1){1}}}
\end{picture}
}\hspace*{3pt}}} 
} 
\LetLtxMacro{\DIFOaddbegin}{\DIFaddbegin} 
\LetLtxMacro{\DIFOaddend}{\DIFaddend} 
\LetLtxMacro{\DIFOdelbegin}{\DIFdelbegin} 
\LetLtxMacro{\DIFOdelend}{\DIFdelend} 
\DeclareRobustCommand{\DIFaddbegin}{\DIFOaddbegin \let\includegraphics\DIFaddincludegraphics} 
\DeclareRobustCommand{\DIFaddend}{\DIFOaddend \let\includegraphics\DIFOincludegraphics} 
\DeclareRobustCommand{\DIFdelbegin}{\DIFOdelbegin \let\includegraphics\DIFdelincludegraphics} 
\DeclareRobustCommand{\DIFdelend}{\DIFOaddend \let\includegraphics\DIFOincludegraphics} 
\LetLtxMacro{\DIFOaddbeginFL}{\DIFaddbeginFL} 
\LetLtxMacro{\DIFOaddendFL}{\DIFaddendFL} 
\LetLtxMacro{\DIFOdelbeginFL}{\DIFdelbeginFL} 
\LetLtxMacro{\DIFOdelendFL}{\DIFdelendFL} 
\DeclareRobustCommand{\DIFaddbeginFL}{\DIFOaddbeginFL \let\includegraphics\DIFaddincludegraphics} 
\DeclareRobustCommand{\DIFaddendFL}{\DIFOaddendFL \let\includegraphics\DIFOincludegraphics} 
\DeclareRobustCommand{\DIFdelbeginFL}{\DIFOdelbeginFL \let\includegraphics\DIFdelincludegraphics} 
\DeclareRobustCommand{\DIFdelendFL}{\DIFOaddendFL \let\includegraphics\DIFOincludegraphics} 

\usepackage{xcolor}
\usepackage{framed}
\definecolor{shadecolor}{rgb}{255,255,0}

\begin{document}

\title{
High Precision Control and Deep Learning-based Corn Stand Counting Algorithms for Agricultural Robot
}


\author{Zhongzhong Zhang \and Erkan Kayacan \and Benjamin Thompson \and Girish Chowdhary
}


\institute{Z. Zhang, B. Thompson and G. Chowdhary \at
 Department of Agricultural and Biological Engineering \\ University of Illinois at Urbana-Champaign. \\
              \email{\{zzhang52, girishc\}@illinois.edu}.        \\
             E. Kayacan  \at 
              School of Mechanical \& Mining Engineering \\
             University of Queensland \\
              \email{e.kayacan@uq.edu.au}    \\
             Z. Zhang and E. Kayacan: Co-primary authors.
}

\date{Received: date / Accepted: date}

\maketitle

\begin{abstract}

This paper presents high precision control and deep learning-based corn stand counting algorithms for a low-cost, ultra-compact 3D printed and autonomous field robot for agricultural operations. Currently, plant traits, such as emergence rate, biomass, vigor, and stand counting, are measured manually. This is highly labor-intensive and prone to errors. The robot, termed TerraSentia, is designed to automate the measurement of plant traits for efficient phenotyping as an alternative to manual measurements. In this paper, we formulate a Nonlinear Moving Horizon Estimator (NMHE) that identifies key terrain parameters using onboard robot sensors and a learning-based Nonlinear Model Predictive Control (NMPC) that ensures high precision path tracking in the presence of unknown wheel-terrain interaction. Moreover, we develop a machine vision algorithm designed to enable an ultra-compact ground robot to count corn stands by driving through the fields autonomously. The algorithm leverages a deep network to detect corn plants in images, and a visual tracking model to re-identify detected objects at different time steps. We collected data from 53 corn plots in various fields for corn plants around 14 days after emergence (stage V3 - V4). The robot predictions have agreed well with the ground truth with $C_{robot}=1.02 \times C_{human}-0.86$ and a correlation coefficient $R=0.96$. The mean relative error given by the algorithm is $-3.78\%$, and the standard deviation is $6.76\%$. These results indicate a first and significant step towards autonomous robot-based real-time phenotyping using low-cost, ultra-compact ground robots for corn and potentially other crops.

\keywords{Machine learning \and Deep learning \and High precision control \and Corn stand counting \and Field robot \and Agricultural robotics.}
\end{abstract}

\section{Introduction}
Phenotypic traits are measured manually by field technicians in the field to determine physical differences between plant genotype and the influence of environmental conditions. Frequent and accurate measurement of these phenotypic traits can be utilized to breed improved crops that have more nutritional value, yield, and resilience to weather anomalies. However, manual phenotyping is expensive due to its labor-intensive nature, and prone to human measurement errors. This has led to the so-called phenotyping bottleneck preventing rapid advances in plant breeding \cite{furbank2011phenomics, araus2014field}. On the other hand, these crop characteristics can also be measured by exteroceptive sensors such as hyperspectral cameras, stereo cameras, thermal cameras, and lidar.  These sensors can be integrated into a mobile robotic scanning and data processing system. Such robotic phenotyping systems have a tremendous potential to relieve the phenotyping bottleneck by significantly reducing labor costs and eliminating errors due to human-subjective information \cite{Young2018}.

\begin{figure}[t!]
\center
  \includegraphics[width=1\columnwidth]{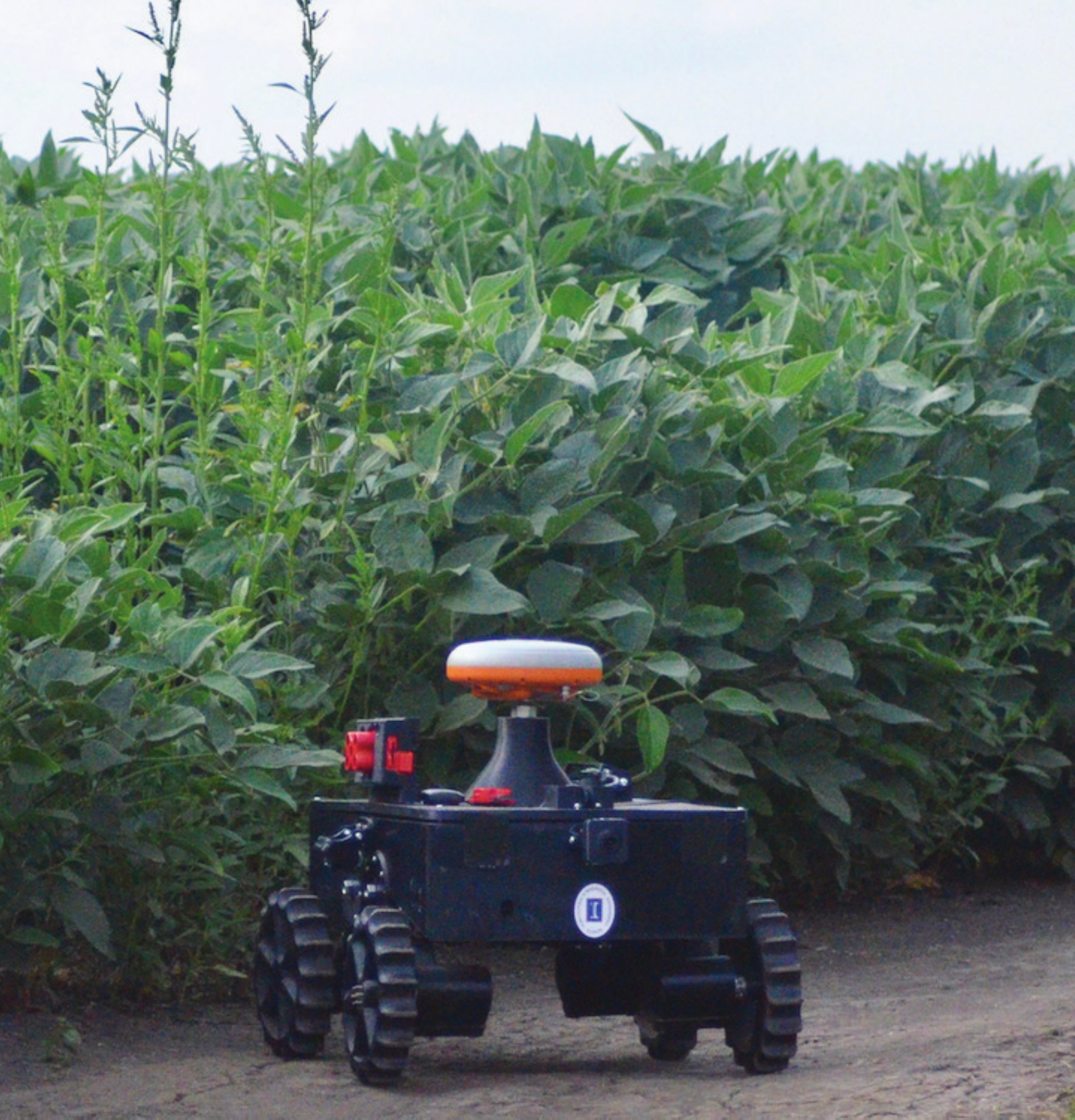} 
  \caption{The ultra-compact 3D printed field robot, termed TerraSentia, in soybean breeding plots in Energy Farm, Urbana, IL, USA.
}\label{fig_real}
\end{figure}

Unmanned robots have the potential to capture extremely detailed data of plants without the need of an expert operator. A four-wheeled autonomous rover with variable chassis clearance and width is equipped with 3D time-of-flight cameras, spectral imaging, light curtain, and a laser distance sensor in \cite{Henten2009, Biber2012}. The rover has many promising features such as increased battery power, battery capacity, and continuous drive torque; however, its wide tracks and limited clearance are serious limitations for plant phenotyping. A new architecture consisting of two mobile robotic platforms, i.e., a phenotyping rover and a mobile observation tower, has been introduced for high-throughput field phenotyping \cite{Shafiekhani2017}. The phenotyping rover has been built on a Husky A-200 from Clearpath Robotics and collects data from individual plants, while the observation tower identifies specific plants for further inspections. Thus, it eliminates the need for air vehicles. However, a Husky A-200 is too wide (66 centimeters) and heavy (50 kilograms). A ground-based agricultural robot for high-throughput crop phenotyping was developed in \cite{7989418}. It is capable of autonomously navigating below the canopy of row crops and deploying a manipulator to measure plant stalk strength and gathering phenotypic data.  However, like a Husky A-200, it is too wide (56 centimeters) and heavy (140 kilograms). TerraSentia designed in this paper is much narrower (30.5 centimeters) and light (6.6 kilograms) when compared to the aforementioned agricultural robots, as shown in Fig. \ref{fig_real}. Thus, it easily fits between crop-rows and does not compact the soil, which is highly desirable for improving yield.

\subsection{Contributions}

This paper describes an ultra-compact (30 centimeters wide), ultra-light (6.6 kilograms), very low-cost autonomous field-phenotyping robot that can navigate in a variety of field conditions to overcome the aforementioned limitations with existing field-based phenotyping systems. A wide variety of systems, e.g., tractors, unmanned aerial vehicles, rovers, and gantry systems, were developed for field-based phenotyping. However, these systems have many limitations, including high operational and maintenance costs, low coverage area, safety, logistically difficult to maintain internal combustion engines, or the need for experienced operators \cite{8409989}. On the other hand, the ultralight robot described here mitigates many of these challenges, yet, it leads to a uniquely challenging control problem in uneven and unstructured terrain in crop fields. The difficulties in control arise from complex and unknown wheel-terrain interaction (described in detail in Section \ref{sec:sysmodel}) and wheel slip. In this paper, a Nonlinear Moving Horizon Estimator (NMHE) that identifies key terrain parameters using onboard robot sensors, and a learning-based Nonlinear Model Predictive Control (NMPC) that enables highly accurate tracking are developed to ensure autonomous high-precision robot mobility in off-road terrain.
The other significant contribution of this study is a novel real-time machine vision algorithm that can estimate plant stand count from image sequences obtained from a side-facing camera on an ultra-compact ground robot. This algorithm is demonstrated on the challenging problem of counting corn (Zea mays or Maize) plants in field-conditions; however, the algorithm may also be re-purposed to count other plants, including Sorghum, Wheat, Soybean, or vegetables. The algorithm leverages a cutting-edge convolution neural network architecture that runs efficiently on mobile platforms. Deep learning methods have been shown capable of recognizing complex structures and features in the presence of heavy noise. Today, deep neural networks are approaching the human level at image recognition on Internet data \cite{he2016deep, krizhevsky2012imagenet, szegedy2015going, simonyan2014very}. In particular, deep learning models have been successfully adopted in several agricultural applications. Lu et al.~\cite{lu2017tasselnet} used neural networks (TasselNet) to count maize tassels from overhead images. The network learns a regression function that maps the input RGB image to a density map whose integral gives the count of objects. The results were evaluated on 361 field images over six years. Their model reported state-of-the-art performance and a mean absolute error of 6.6. In addition, Rahnemoonfar and Sheppard~\cite{rahnemoonfar2017deep} trained a neural network with only synthetic data generated by randomly drawing red circles on a green-brown background to mimic the appearance of tomatoes. Nonetheless, the network showed $91\%$ average accuracy when tested on real images. More recently, Chen et al.~\cite{chen2017counting} presented a three-stage model to count oranges and apples. The model consists of a fully convolutional network that segments fruits from the background, a second network that counts the fruits in a segmentation blob, and finally, a linear regressor that relates the count in each blob to the total count in the image. However, these studies still target only images, while counting vegetables and fruits in a video remains a challenge. In this study, we present a robust stand-counting algorithm where the only sensor needed is a low-cost ($< \$ 30$) RGB camera. Our extensive field trials show that the detection is robust against noises such as corn leaves, weeds, varying lighting conditions, and residues from the previous year. Our system achieves high accuracy and reliability throughout the growing season. In essence, this contribution opens the door to real-time robotic phenotyping in breeding plots and production fields. Robotic phenotyping is a welcome contribution since it has the potential to overcome the phenotyping bottleneck \cite{araus2014field, furbank2011phenomics} that has slowed progress in breeding better crops.

Elements of this work appeared in a Robotics Science and Systems conference paper \cite{KayacanRSS}. Significant additions include details of the design of the 3D printed field robot and also a new deep-learning-based corn stand counting algorithm.

\subsection{Organization of the paper}

This paper is organized as follows: The design of the 3D printed field robot is given in Section \ref{sec_design}. The developed NMHE-NMPC framework for high precision control of the ultra-compact mobile robot in off-road terrain is described in Section \ref{sec_control}. The developed real-time deep learning-based machine vision algorithm for corn stand counting is given in Section \ref{sec_mac}. Field results are given in Section \ref{sec_exp_res}, and Section \ref{sec_conc} concludes the paper.


\section{Design of 3D Printed Field Robot}\label{sec_design}

In this paper, we develop a novel robot that is constructed mostly out of 3-D printed construction. This leads to an incredibly lightweight robot (6.5 kg), yet, the robot has been proven to be structurally resilient to field conditions during an entire season of heavy operation in Corn, Sorghum, and Soybean farms in Illinois. We posit that this robot is an example of the potential of additive manufacturing (3-D printing) in creating a new class of agricultural equipment, which is lightweight, easier to manage, safer to operate in general, and leads to lower ownership cost.

\subsection{Mechanical Design}
\begin{figure}[t!]
  \centering
  \includegraphics[width=1\columnwidth]{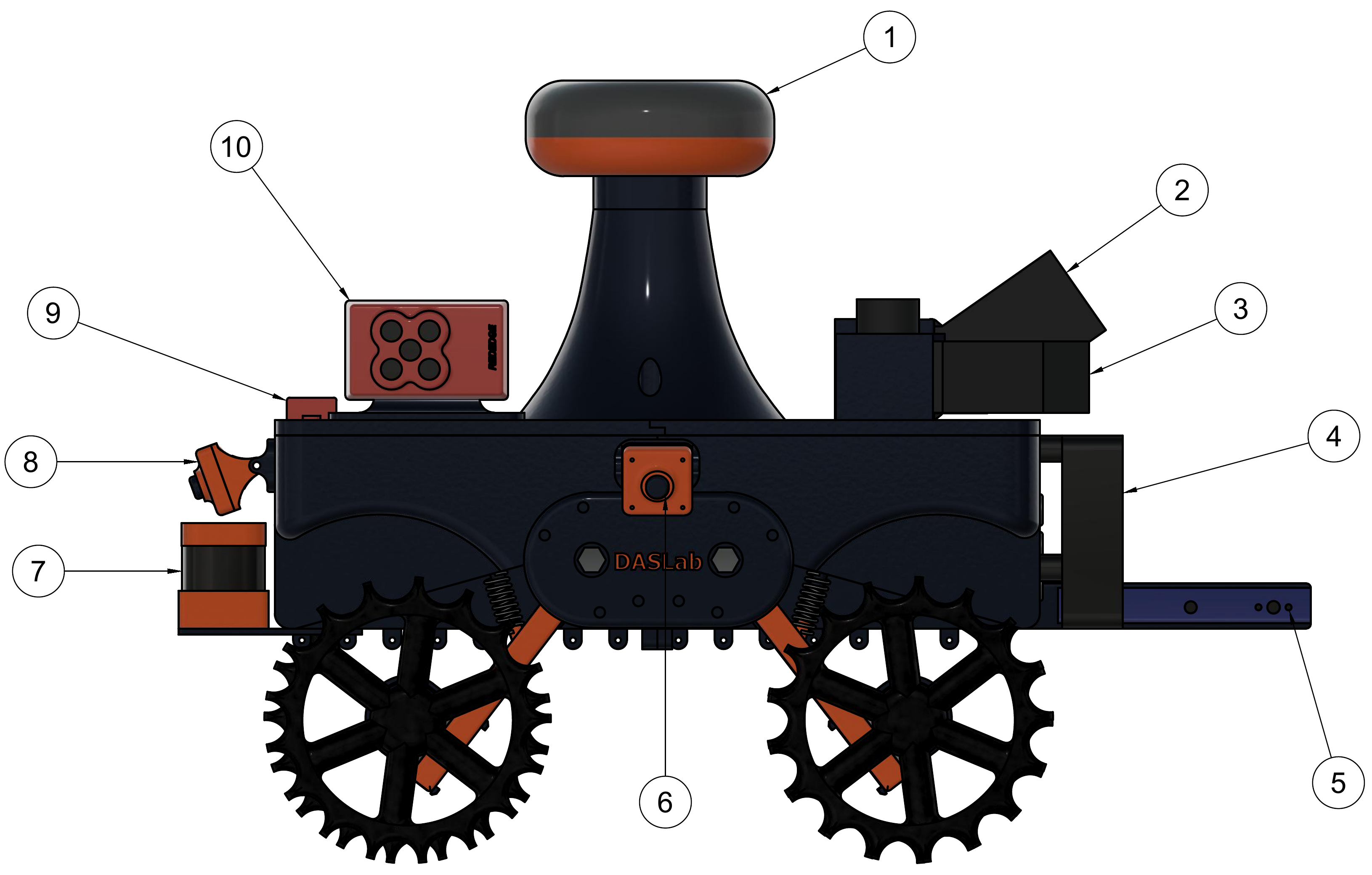} 
  \caption{CAD drawing of the ultra-compact 3D printed robot with a suite of sensors. 1. GNSS antenna, 2. Bayspec hyperspectral sensor, 3. Bayspec hyperspectral sensor (sideward facing), 4. Radiator for the liquid cooling system, 5. Mount for 3d Sensor Intel RealSense, 6. Embedded visual sensor, 7. LIDAR sensor, 8. Embedded visual sensor, 9. GNSS mount for RedEdge multispectral sensor, 10. RedEdge multispectral sensor.}\label{fig_diagram}
\end{figure}

By leveraging 3D printing, our aim is to produce a robot that is much lighter by using less metal in the construction of the robot. A Computer-Aided Design (CAD) drawing of the robot with a suite of sensors attached is depicted in Fig. \ref{fig_diagram}. The robot has been explicitly designed to be ultralight (less than 7 kg) and compact. The ultralight design requires significant thought in the selection of material, the construction of the material, and the structural design. Conventional methods for manufacturing agricultural equipment often use metal; however, metal is heavy and expensive. As a stark contrast, our robot is built through novel mechanisms of additive manufacturing, a method of manufacturing which heats and extrudes a thermoplastic filament to produce a three-dimensional object. This novel mechanism has not yet been used to design agricultural robots. Additive manufacturing creates complex designs by layering material, as opposed to traditional metalworking, or traditional injection molding of plastics. By leveraging the capabilities of additive manufacturing, the aim is to develop novel designs that minimize the weight while maintaining sufficient structural rigidity. These design features allow us to design a robot that optimally distributes stress and controls the infill percentage allowing for increased strength and durability while being lightweight. Mechanical parts are denser and stronger, allowing non-load bearing components to be much lighter. 

Another challenge is ensuring the ground clearance of the robot, which is sufficiently high to enable traversing complex terrain. The challenge here is with placement. Specifically, motors need to be placed close enough to the robot wheels to minimize the loss in transmission and to keep it mechanically simple; meanwhile, the placement of the motors must allow the robot to operate through continued use in harsh field conditions. This is achieved by creating novel wheel mechanisms that ensure minimal stress on the motors, as shown in Fig. \ref{fig_drivetrain}.

\subsection{Wheel Design}

The appropriate wheel design is critical in ensuring that the robot can navigate over wet and muddy terrains. Furthermore, wheel design affects the robot-plant interaction if the robot does drive over young plants. 3D printing enables the rapid testing of many different tread designs to determine optimal tread patterns for the environment. We use flexible filament, which has similar properties to rubber while still being able to optimize the design of the tread. The wheels use spade-like arrangements that are designed to provide traction on loose soils while minimizing the contact area. As opposed to tracked robots, this wheel design has significant advantages: 
\begin{itemize}
\item The design does not lead to a large area subject to pressure and force as the robot moves; instead, the wheel design is limited to a small contact area. Thus, if the robot does drive over leaves of older plants, it is not likely to drag in the whole plant under its body;
\item The design is much simpler to manufacture and operate in the field; 
\item The design is modular, in the sense that each wheel can be replaced if it breaks, instead of having to replace the whole track. 
\end{itemize}

\begin{figure}[t!]
  \centering
  \includegraphics[width=1\columnwidth]{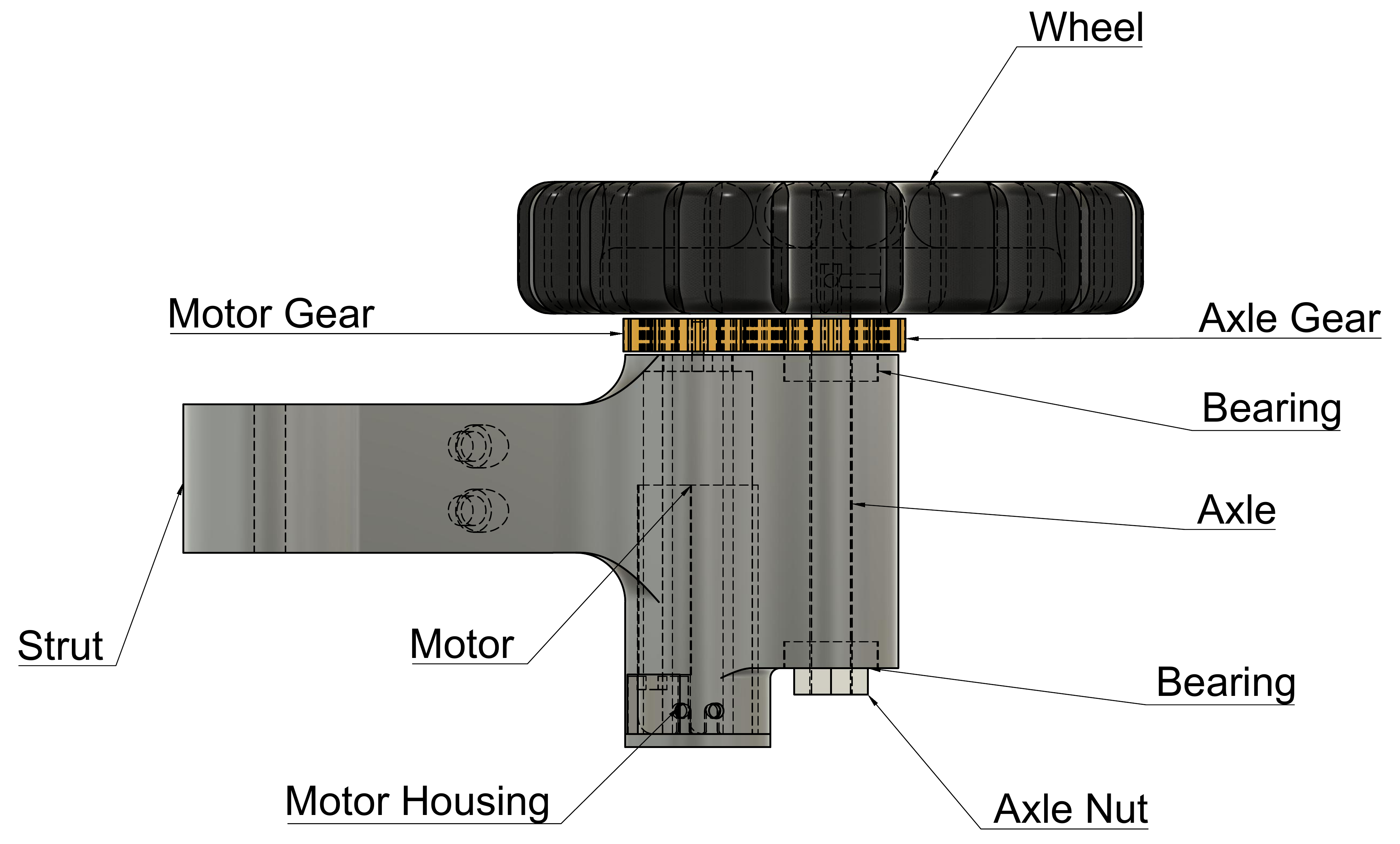} 
  \caption{CAD drawing of the drivetrain.}\label{fig_drivetrain}
\end{figure}

The driving mechanism, as illustrated in Fig. \ref{fig_drivetrain} includes motors mounted near each wheel to enable four independently driven wheels without the need to distribute power from a central power unit. Such a driving mechanism is distinct from existing equipment and vehicles, which utilize a single power plant to transmit power to different wheels. The four-wheel-drive mechanism enables the robot to turn by varying the speeds of independent wheels and is a much simpler mechanism because it does not require complex rack-and-pinion or other similar mechanisms for driving. Another feature of the wheel and mount design is the embedded suspension without having to increase the size of the robot. The suspensions are embedded between the wheel mount and the chassis. The broad chassis provides a simple mechanism that can easily handle bumpy agricultural fields.

\subsection{Hardware}

The platform used in this study is a four-wheel mobile robot, which is \SI{30}{cm} tall $\times$ \SI{50}{cm} long $\times$ \SI{35}{cm} wide, with a \SI{15}{cm} ground clearance, and weighs \SI{6.5}{kg}. Such a compact and lightweight design allows the robot to easily traverse between crop rows of typical row spacing for corn. The robot is powered by four Lithium-Ion batteries that offer up to \SI{8}{\hour} duration.
\begin{figure}[t!]
  \centering
  \includegraphics[width=1\columnwidth]{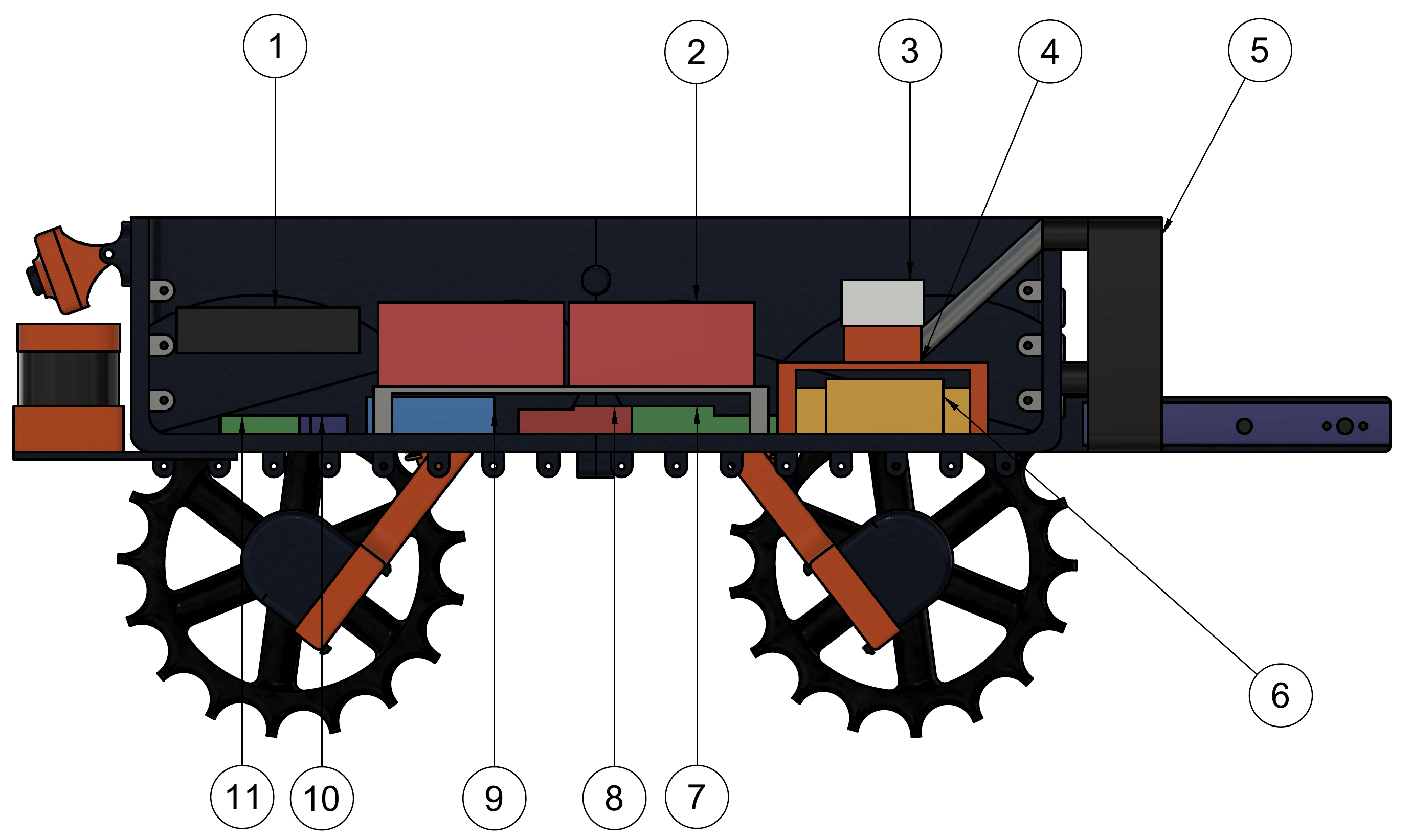} 
    \caption{Interior of the ultra-compact 3D printed robot. 1. Raspberry Pi, 2. Lithium-Ion Batteries, 3. Tegra, 4. Heat sink, 5. Cooling Fan, 6. Kangaroo/Sabertooth, 7. A regulator, 8. 3-axis gyroscope, 9. Breadboard, 10. Raspberry Pi, 11. 3-axis accelerometer. }\label{fig_diagram_interior}
\end{figure}

One Global Navigation Satellite System (GNSS) antenna has been mounted straight up the center of the 3D printed robot, and the dual-frequency GPS-capable real-time kinematic differential GNSS module (Piksi Multi, Swift Navigation, USA) has been used to acquire centimeter-level accurate positional information at a rate of 5 Hz. Another antenna and module have been used as a portable base station and has transmitted differential corrections.  The placement of hardware is illustrated in Fig. \ref{fig_diagram_interior}. A 3-axis gyroscope (STMicroelectronics L3G4200D) has been used to obtain yaw rate measurement with an accuracy of 2 degrees per second at a rate of 5 Hz. There are four brushed 12V DC motors with a 131.25:1 metal gearbox (Pololu Corporation, USA), which are capable of driving an attached wheel at 80 revolutions per minute. A two-channel Hall-effect encoder (Pololu Corporation, USA) for each DC motor is attached to measure velocities of the wheels. The Sabertooth motor controller (Dimension engineering, USA) is a two-channel motor driver that uses digital control signals to drive two motors per channel (left and right channel) and has a nominal supply current of 12 A per channel. The Kangaroo x2 motion controller (Dimension engineering, USA) is a two-channel self-tuning proportional-integral-derivative controller that uses feedback from the encoders to maintain desired linear and angular robot velocity commands. An onboard computer (1.2 GHz, 64bit, quad-core Raspberry Pi 3 Model B CPU) acquires measurements from all available sensors and sends desired control signals (e.g., desired linear and angular velocities) to the Kangaroo x2 motion controller in the form of two Pulse-width modulation signals.

\begin{figure}[t!]
\centering
    \subfigure[The robot in a cornfield between two rows]{
        \includegraphics[width=0.7\linewidth]{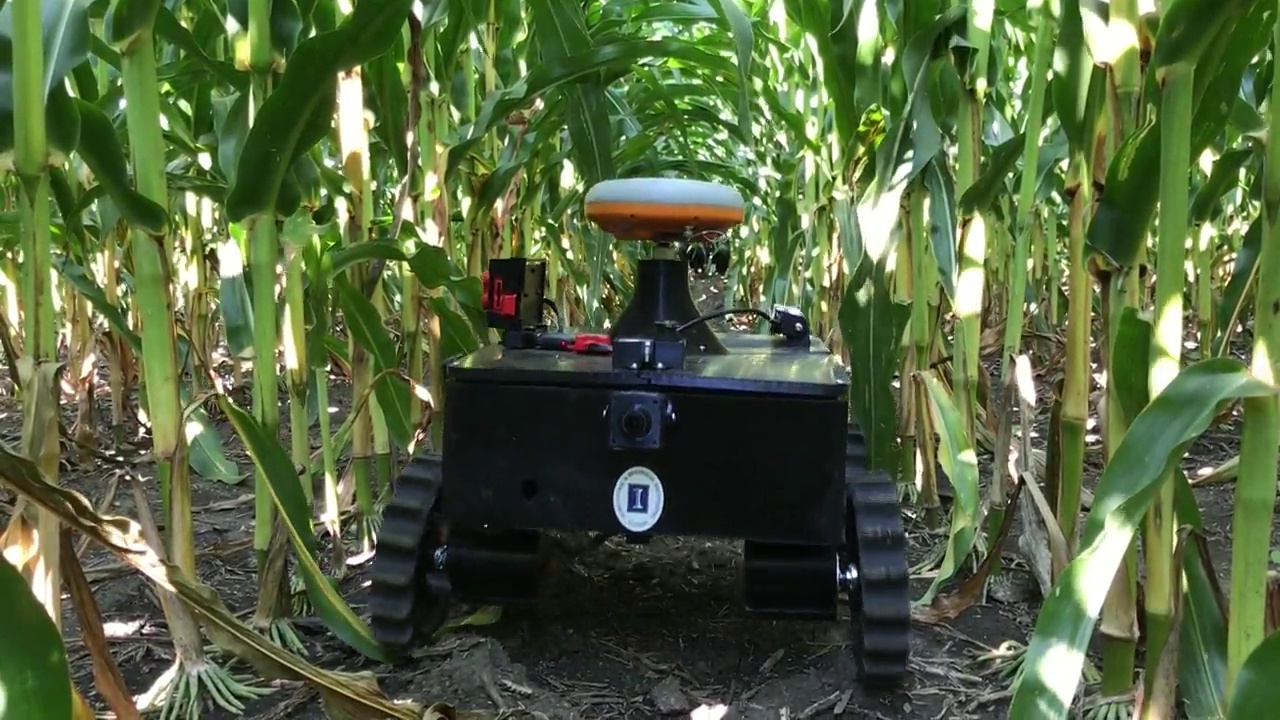}
    } 
    \subfigure[Posterior view of the robot]{
        \includegraphics[width=0.9\linewidth]{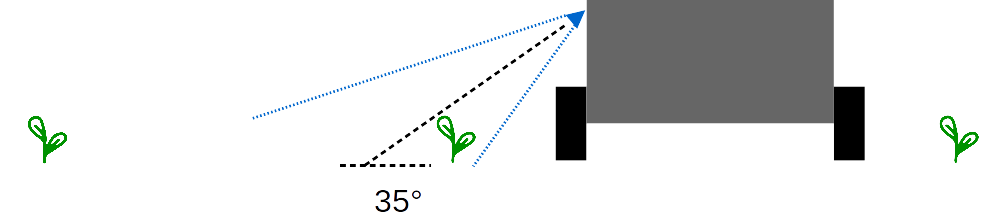}
    }
        \subfigure[Superior view of the robot]{
        \includegraphics[width=0.8\linewidth]{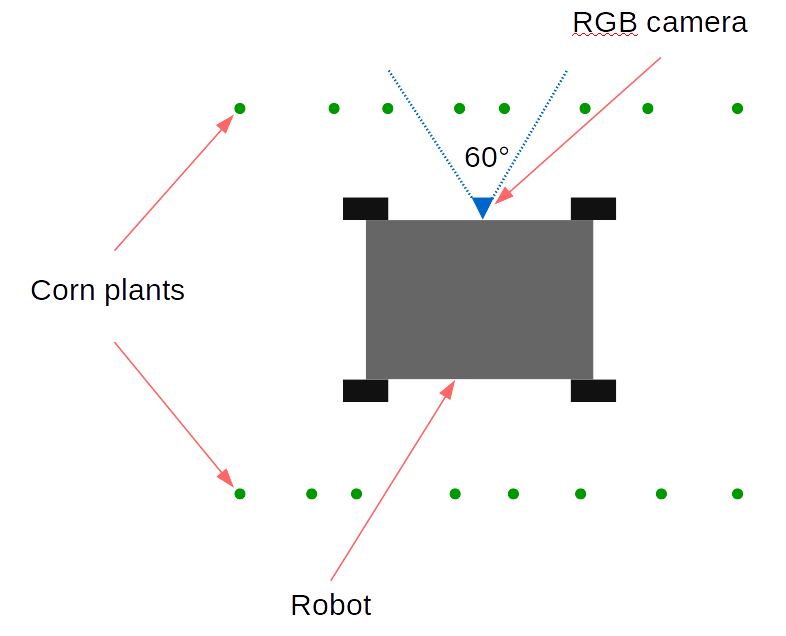}
    } 
    \caption{Illustrations of the ground robot. An RGB camera mounted on the side of a robot records video as the robot traverses between two corn rows. The camera has a field of view of \ang{60} and points downward at \ang{35}.}
    \label{fig:robot}
\end{figure}

Images are recorded with an RGB digital camera (ELP USBFHD01M, USA) mounted on the side of the robot chassis. The field of view of the camera is \ang{60}. The number of corn plants captured in the image depends on the distance between the camera and plant row, as well as the spacing between adjacent plants. In a 75-cm row, for instance, two to three corn plants commonly appear in the image. The camera points down at an angle of \ang{35} to avoid observing corn rows far away. An illustration of the data acquisition system set-up is shown in Fig. \ref{fig:robot}. The resolution of the camera is $640 \times 480$, and it records at \num{30} frames per second. The camera has a USB 2.0 interface that connects to a Jetson TX2 (NVIDIA, USA), an embedded module for fast and efficient deep neural network inference. The module houses \SI{8}{\giga \byte} memory that is shared between CPU and GPU and can process image frames captured by the camera in real-time.

\section{High Precision Control Algorithm}\label{sec_control}

This section presents a high precision control algorithm in which Nonlinear Moving Horizon Estimator (NMHE) identifies key terrain parameters using onboard robot sensors, and a learning-based Nonlinear Model Predictive Control (NMPC) ensures high precision path tracking in the presence of unknown wheel-terrain interaction.

\subsection{System Model}\label{sec:sysmodel}

The effect of terrain characteristics on robot performance and wheel requires special attention in field-robot design because the wheels are the unique connections between the ground and robot, and nearly all forces and moments applied to the robot are transmitted through the wheels \cite{1339393, 4840546}. Due to the wheel-terrain interaction dynamics, soil characteristic plays a vital role in determining vehicle speed and steering, which in turn are utilized for developing traction control algorithms. The knowledge of soil parameters of unknown terrain is then advantageous for improving vehicle performance \cite{6606388, 7759528, 7487413, KayacanJFR, Mehndiratta2019}. In particular, for lightweight robots meant for persistent field use, such as TerraSentia robot described here, careful design of control algorithms is necessary to ensure that the robot does not tread over plants due to loss of precision due to unaccounted for wheel-soil interaction.

A system model including traction parameters is developed theoretically and used to establish an effective control law for the 3D printed field robot traveling on rough terrain in this paper. An adaptive nonlinear kinematic model is derived for the ultra-compact 3D printed field robot as an extension of the traditional kinematic model as follows: 
\begin{eqnarray}\label{eq_systemmodel}
 \left[
  \begin{array}{c}
   \dot{x} \\ \dot{y} \\ \dot{\theta} 
  \end{array}
  \right] = \left[
  \begin{array}{c}
   \mu v \cos{\theta}\\  \mu v \sin{\theta} \\ \kappa \omega 
  \end{array}
  \right]
\end{eqnarray}
where $x$ and $y$ are the position of the field robot, $\theta$ is the yaw angle, $v$ is the wheel speed, $\omega$ is the yaw rate, and $\mu$ and $\kappa$ are the traction parameters. The difference between the traditional and developed model above is two slowly changing traction parameters. These parameters provide us a learning mechanism that can be used to minimize deviations between the real-time system and ultra-compact 3D printed field robot with an online parameter estimator. It is noted that they must be between zero and one. If the traction parameters are equal to 1, i.e., $\mu=\kappa=1$, there exist no longitudinal or side (lateral) slips. The percentages of the longitudinal and side slips can be respectively found as $1-\mu$ and $1-\kappa$. It is assumed that only a fraction of the realized speed and yaw rate translates into actual vehicle motion. These fractions are given by $\mu$ and $\kappa$, and determine the effective speed $\mu v$ and yaw rate $\kappa \omega$. To avoid bias, these parameters should be estimated along with the full system state in each iteration based on a number of past measurements. 

In this paper, the nonlinear system and measurement models are represented by the following equations:
\begin{eqnarray}\label{eq_systemmodel}
\dot{\xi}(t) = f \Big( \xi(t),u(t),p(t) \Big) \\ 
z(t) = h \Big( \xi(t),u(t),p(t) \Big)
\end{eqnarray} 
where $\xi$ $\in$ $\mathbb{R}^{n_{\xi}}$ is the state vector, $u$ $\in$ $\mathbb{R}^{n_{u}}$ is the control input, $p$ $\in$ $\mathbb{R}^{n_{p}}$ is the system parameter vector, $z$ $\in$ $\mathbb{R}^{n_{z}}$ is the measured output, $f(\cdot,\cdot,\cdot): \mathbb{R}^{n_{\xi}+n_{u}+n_{p}}  \longrightarrow \mathbb{R}^{n_{\xi}} $ is the continuously differentiable state update function and $f(0,0,p)=0 \; \forall t$, and $h: R^{n_{\xi}+n_{u}+n_{p}} \longrightarrow R^{n_{z}}$ is the measurement function. The derivative of $\xi$ with respect to $t$ is denoted by $\dot{\xi}$ $\in$ $\mathbb{R}^{n_{\xi}}$. The state, parameter, input and output vectors are respectively denoted as follows:
\begin{eqnarray}
\label{eq_state}
\xi & = & \left[
  \begin{array}{ccc}
   x & y & \theta 
  \end{array}
  \right]^{T} \\
p &=& \left[
  \begin{array}{ccc}
   v &\mu & \kappa 
  \end{array}
  \right]^{T}   \nonumber  \\ 
u & = &
  \begin{array}{c}
   \omega 
  \end{array} \\ 
  z &=& \left[
  \begin{array}{cccc}
   x & y & v & \omega 
  \end{array}
  \right]^{T}
\end{eqnarray}

\subsection{Nonlinear Moving Horizon Estimation}

Although model-based controllers need full state and parameter information to generate a control signal, the number of sensors is less than the number of measurable states and parameters in practice. Therefore, state estimators are required to estimate immeasurable states and parameters. Extended Kalman filter is the most well-known state-estimation method for nonlinear systems. However, EKFs are not capable of dealing with constrained nonlinear systems \cite{Haseltine2005}. As can be seen in \eqref{eq_systemmodel}, the traction parameter estimates play a vital role for the ultra-compact 3D printed robot, and there exist constraints on these parameters, which makes Extended Kalman filter inconvenient for our system. Nonlinear Moving Horizon Estimation (NMHE) approach has a capability of dealing with constraints on state and parameter and is formulated for the ultra-compact 3D printed robot as follows:

\begin{equation}
 \begin{aligned}
 & \underset{\xi(t),p,u(t)}{\text{min}}
 & &  \frac{1}{2} \bigg\{  \left\|
  \begin{array}{c}
    \hat{\xi} (t_{k-N_{e}+1}) - \xi (t_{k-N_{e}+1})  \\
    \hat{p} - p
 \end{array}
 \right\| ^{2}_{H_N}  \\
 &&&+ \sum_{i=k-N_{e}+1}^{k}  \| z_m(t_i) - z (t_i) \|^{2}_{H_k} \bigg\}  \\
 & \text{ s. t. }
 &&  \dot{\xi}(t) = f \big(\xi(t),u(t),p \big) \\
 &&&  z(t) = h \big(\xi(t),u(t),p \big) \\
 &&& 0 \leq \mu \leq 1  \\
 &&& 0 \leq \kappa \leq 1 \qquad \qquad \forall t \in [t_{k-N_{e}},  t_k]\\
  \end{aligned}
    \label{eq_nmhe}
\end{equation}
where the deviations of the state and parameter estimates before the estimation horizon are minimized by a symmetric positive semi-definite weighting matrix $H_N$, and the deviations of the measured and system outputs in the estimation horizon are minimized by a symmetric positive semi-definite weighting matrix $H_k$ \cite{Ferreau}. The estimation horizon is represented by $N_{e}$, and lower and upper bounds on the traction parameters $\mu$ $\kappa$ are respectively defined as 0 and 1. The objective function in the NMHE formulation consists of two parts: the arrival and quadratic costs. The arrival cost stands for the early measurements $t=[t_{0, k-N_{e}}]$, and the quadratic cost stands for the recent measurements $t=[t_{k-N_{e}+1,  k}]$.

The measurements have been perturbed by Gaussian noise with standard deviation of $\sigma_{x} = \sigma_{y} = 0.03$ m, $\sigma_{\omega} = 0.0175$ $rad/s$, $\sigma_{v} = 0.05$ $m/s$ based on experimental analysis. Therefore, the following weighting matrices $H_{k}$ and $H_{N}$ are used in the NMHE:
\begin{eqnarray}\label{eq_mhe_weightingmatrices}
H_{k} & = & diag(\sigma_{x}^{2},\sigma_{y}^{2},\sigma_{v}^{2}, \sigma_{\omega}^{2})^{-1} \nonumber \\
        & = & diag(0.03^{2},0.03^{2},0.5^{2},0.35^{2})^{-1} \nonumber \\
H_{N} & = & diag(x^{2}, y^{2}, \theta^{2}, v^{2}, \mu^{2}, \kappa^{2})^{-1} \nonumber \\
      & = & diag(10.0^{2}, 10.0^{2}, 0.1^{2}, 1.0^{2}, 0.25^{2}, 0.25^{2})^{-1}
\end{eqnarray}

The inputs to the NMHE algorithm are the position values coming from the global navigation satellite system, the velocity values coming from the encoders mounted on the DC motors and the yaw rate values coming from the gyro. The outputs of NMHE are the position in x- and y-coordinate system, the yaw angle, the wheel velocity and the traction coefficients. The estimated values are then fed to the NMPC.

\subsection{Nonlinear Model Predictive Control}

Nonlinear Model Predictive Control (NMPC) approach is the most well known control method for system with fast dynamics. The reason is that NMPC has the capability of dealing with hard constraints on state and input, and online optimization allows updating cost, model parameters, constraints \cite{Kayacan2018}. The following finite horizon optimal control formulation for the ultra-compact 3D printed robot is solved to obtain the current control action by using current states  and parameters of the system as initial state: 
\begin{equation}
 \begin{aligned}
 & \underset{\xi(t), u(t)} {\text{min}}
 & & \frac{1}{2} \bigg\{  \Big\{ \sum_{i=k+1}^{k+N_{c}-1}  \| \xi_{r} (t_{i}) - \xi (t_{i}) \|^{2}_{Q_{k}} + \|  u (t_{i}) \|^{2}_{R}  \Big\} \\
 &&& \qquad + \| \xi_{r} (t_{k+N_{c}}) - \xi (t_{k+N_{c}}) \|^{2}_{Q_{N}}  \bigg\} \\
 & \text{s. t.}
 && \xi(t_k) = \hat{\xi} (t_k) \\
  &&& p = \hat{p} (t_{k-N_{e}+1}) \\
 && & \dot{\xi}(t) = f \big(\xi(t), u(t), p \big) \\
&& & - 0.1 \; \mathrm{rad}/s \leq \omega (t) \leq 0.1 \; \mathrm{rad}/s \quad  t \in [t_{k}, t_{k+N-1}] \\
  \end{aligned}
  \label{eq_nmpc}
\end{equation}
where $Q_{k} \in \mathbb{R}^{n_{\xi} \times n_{\xi}}$, $R \in \mathbb{R}^{n_{u} \times n_{u}}$ and $Q_{N} \in \mathbb{R}^{n_{\xi} \times n_{\xi}}$ are symmetric and positive semi-definite weighting matrices, $\xi_{r}$ is the state reference, $\xi$ and $u$ are the states and inputs, $t_k$ is the current time, $N_{c}$ is the prediction horizon, $\hat{\xi} (t_k)$ is the current state estimates and $\hat{p} (t_{k-N_{e}+1})$ is the current parameter estimates. The first term in the cost function is the stage cost and it is the cost throughout the prediction horizon. The second term in the cost function is the terminal penalty and it is the cost at the end of the prediction horizon. The terminal penalty is stated for stability reasons \cite{Mayne}. The weighting matrices $Q_{k}$, $R$ and $Q_{N}$ are selected as follows:
\begin{equation}
Q_{k} = diag(1,1,1),  \quad R = 1 \quad \textrm{and} \quad Q_{N}  = 10 \times Q_{k} 
\end{equation}
The first element of the optimal control sequence is applied to the system:
\begin{equation}
u(t_{k+1},\xi_{k+1})= u^*(t_{k+1})
\label{U}
\end{equation}
and then the procedure is repeated for future sampling times by shifting prediction horizon for the subsequent time instant. It is important to remark that the control input $u^*(t_{k+1})$ is precisely the same as it would be if all immeasurable states and parameters acquire values equal to their estimates based on the estimation up to current time $t_{k}$ due to the certainty equivalence principle.

The state reference for the 3D printed robot is dynamically changing online and defined as follows:
\begin{eqnarray}\label{eq_mpc_ref}
\xi_{r} = [x_{r},y_{r}, \theta_{r} ]^T
\end{eqnarray}
where $x_{r}$ and $y_{r}$ are the position references, $\omega_{r}$ is the yaw rate reference calculated from the position references as follows:
\begin{eqnarray}\label{eq_mpc_ref_ya}
\theta_{r} = \atantwo{( \dot{y}_{r}, \dot{x}_{r})} + \lambda \pi
\end{eqnarray}
where $\lambda$ describes the desired direction of the 3D printed field robot ( $\lambda=0$ for forward and $\lambda=1$ for backward). 

The yaw rate reference can be calculated from the reference trajectory as the yaw angle reference. However, steady state error may occur in the case of a mismatch between the system model and 3D printed robot \cite{erkan2016acc}. Therefore, the recent measured yaw rate is used as the input reference to penalize the input rate in the objective function in this paper. 

\subsection{Solution Methods}

NMHE and NMPC methods for systems require online solutions of nonlinear least square optimization problems at each sampling time \cite{erkan2016acc}. In this study, a single solution method, consisting of a fusion between multiple shooting and generalized Gauss-Newton methods, has been used to solve both optimization problems. This approach is valid because the formulation of NMPC problem is akin to that NMHE problem. The generalized Gauss-Newton method derived from the classical Newton method was developed for least-squared problems. This method is advantageous because it does not require difficult computations of the second derivatives; however, it is challenging to foreknow the number of iterations to reach a solution of the desired accuracy. To overcome this challenge, the solution proposed in \cite{Diehl2} has been used where the number of Gauss-Newton iterations is restricted to $1$, and the initial value of each optimization problem takes on the value of the previous one intelligently. Hence, this improves the convergence of the Gauss-Newton method.

In this paper, the Gauss-Newton iteration is divided into two parts: preparation and feedback parts.  The preparation parts are executed before the feedback parts, and the feedback part is executed after measurements for NMHE and estimates for NMPC are available. In the preparation part, the system dynamics are integrated with the previous solution, and objectives, constraints, and corresponding sensitives are evaluated. In the feedback part, a single quadratic program is solved with the current measurements for the NMHE and the current estimates for the NMPC. Thus, new estimates for the NMHE and new control signals for the NMPC are obtained. Compared to the classical method, this method minimizes feedback delay and produces similar results with higher computational efficiency.

\section{Corn Stand Counting Algorithm}\label{sec_mac}

This section presents a new algorithm that extends our previous work \cite{KayacanRSS}. Instead of classifying a fixed region of interest, the new algorithm uses an object detection model that can locate objects in the entire image. We then rely on a visual tracking model to establish correspondence in the temporal axis. The overall problem is then decomposed into two sub-tasks: 1) detecting target corn plants in the image, and 2) tracking and identifying the objects across consecutive image frames.

\subsection{Object Detection}

Images captured in the outdoor environment are subject to a wide range of variations such as sunlight, occlusion, and
camera view angle, etc. Additionally, corn plants go through significant changes during the growing season (Figure 6). Deep learning has made tremendous progress in areas that have confounded traditional machine learning for many years \cite{krizhevsky2012imagenet, simonyan2014very}. In particular, convolutional neural networks (CNN or ConvNets) show great promise in computer vision tasks such as image classification \cite{szegedy2015going, he2016deep}, object detection \cite{Ren2015, Liu2015}, and segmentation \cite{long2015fully, he2017mask}. Among various models, Faster R-CNN \cite{Ren2015} achieves state-of-the-art results for object detection on several benchmark datasets. The model takes in an image and outputs. \textit{N} bounding boxes where \textit{N} is the number of objects found in the image, and each bounding box is associated with its size, location, and probability for each class.

\begin{figure*}
	\centering
	\subfigure{\includegraphics[width=0.24\linewidth]{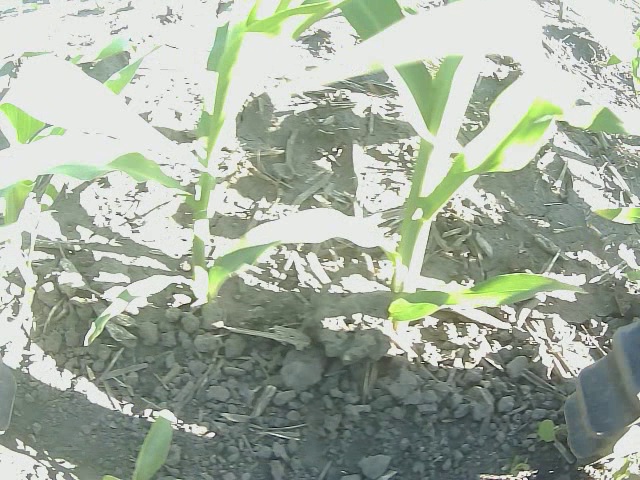}}
	\subfigure{\includegraphics[width=0.24\linewidth]{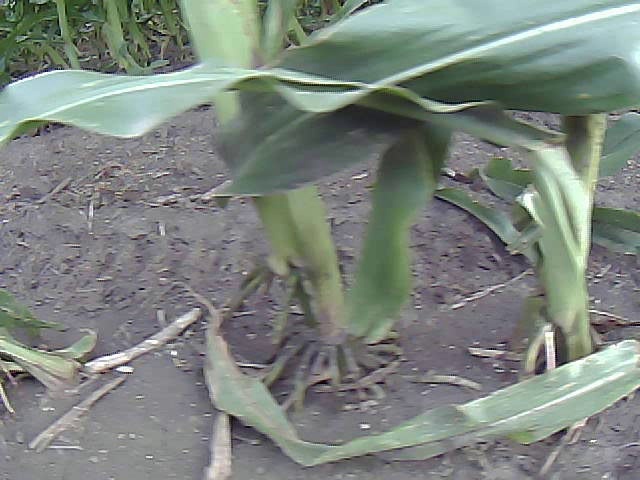}}
	\subfigure{\includegraphics[width=0.24\linewidth]{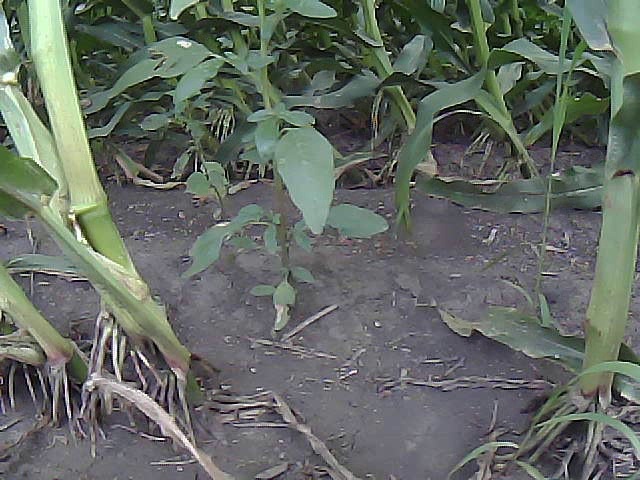}}
	\subfigure{\includegraphics[width=0.24\linewidth]{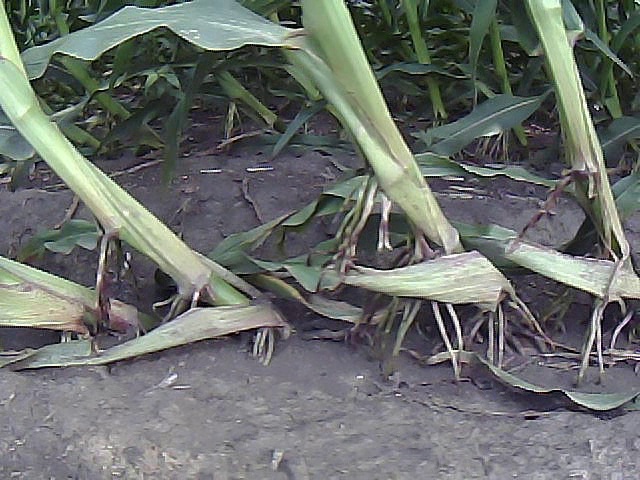}}

	\subfigure{\includegraphics[width=0.24\linewidth]{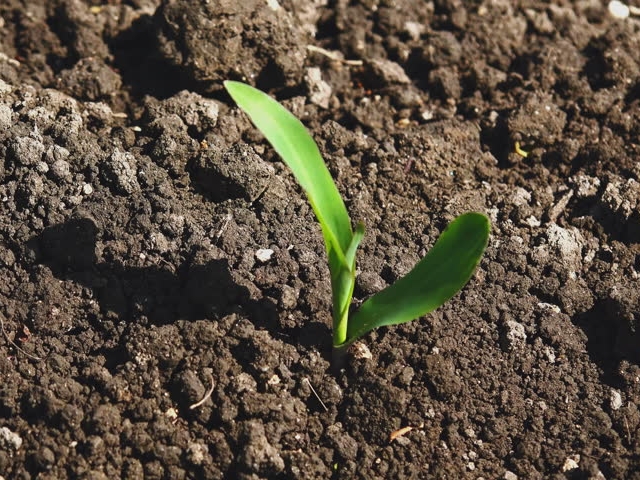}}
	\subfigure{\includegraphics[width=0.24\linewidth]{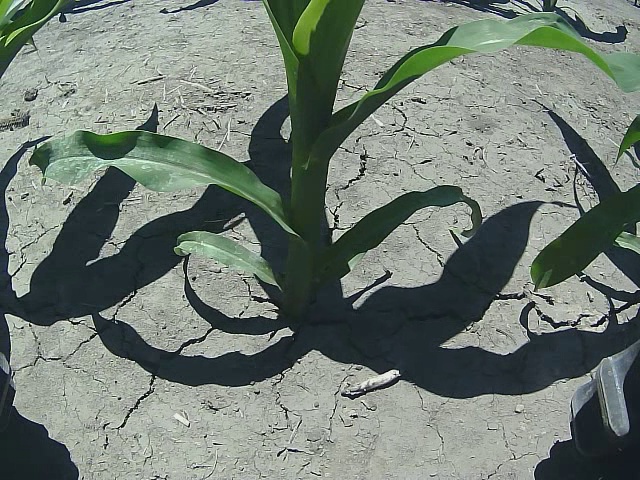}}
	\subfigure{\includegraphics[width=0.24\linewidth]{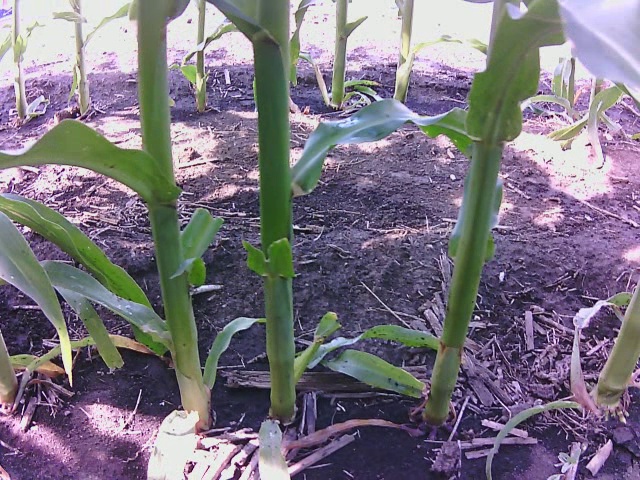}}
	\subfigure{\includegraphics[width=0.24\linewidth]{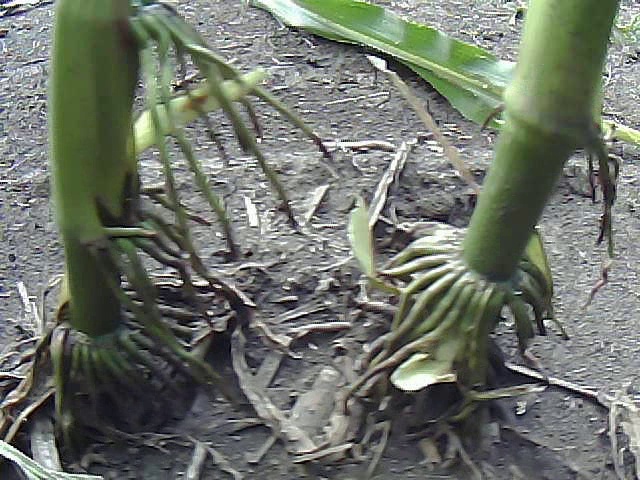}}
	\caption{\label{fig:variations} Sample images demonstrate variations in the environment, field condition, and crop growth stage. Top left to right: Overexposure due to strong sunlight; partial occlusion; heavy weed infestation; and lodging stalk. Bottom left to right: corn plant in V2, V4, VT, and R2.}
\end{figure*}

\subsection{Visual Tracking}

While Faster R-CNN can detect objects in a single image, it has no temporal awareness of the identities of the objects between consecutive frames in a video. Therefore, it is necessary to obtain association across multiple image frames to avoid counting the same object more than once in multiple views. The problem has been widely addressed in studies that attempt to count fruit and crop from image sequences. Two strategies are typically used to establish the association - visual tracking and epipolar projection. \cite{das2015devices} calculated optical flow of image descriptors between sucessive frames to predict the locations of the fruits detected in the previous frames. The predictions are compared with the new detections to assign association. Later works based on visual tracking more or less adopted this method with modifications on the tracking and assignment algorithms. \cite{halstead2018fruit} proposed a 2-stage tracking by detection approach. In stage 1, new detections are paired with active tracks by optimizing intersection over overlap (IoU). In stage 2, the unpaired detections are removed if they have significant overlap with active tracks. The remaining detections are then used to initialize new tracks. \cite{liu2018robust,liu2019monocular} used the Kalman filter to refine the predictions by optical flow, followed by the Hungarian algorithm for assigning the predictions and detections. In addition, the authors applied structure from motion (SfM) to estimating the fruit 3D landmarks to extend the tracking beyond the image plane. The SfM step was reported to further reduce the tracking ambiguity. On the other hand, \cite{stein2016image} developed a tracking algorithm based on epipolar geometry. A ray was projected from the camera focal center of image $I_t$ through the location of the detected fruit in the image plane. The ray was then clipped at a near and a far disctance. Finally, the clipping points were projected to the next image $I_{t+1}$. The projections represent the new locations of the fruits in image $I_{t+1}$. The camera poses were obtained from GPS/INS trajectory data. \cite{liu2019monocular} compared their algorithm based on visual tracking with that of the epipolar projection by Stein et al. The accuracy was comparable without the need for additional instrument. Therefore, we use a visual tracking model to maintain temporal correspondence. 

Suppose the object detection model returns a set of bounding boxes $\pazocal{P}$ and $\pazocal{Q}$ for image $\pazocal{I}_t$ and $\pazocal{I}_{t+1}$, respectively. The task is then formulated as:

\begin{equation}
\pazocal{M} = \argmin \sum_{p\in\pazocal{P}, q\in\pazocal{Q}} c(p, q)
\end{equation}
where $\pazocal{M}$ is the set of matches between $\pazocal{P}$ and $\pazocal{Q}$, and $c(\cdot, \cdot)$ is the cost function for associating two bounding boxes. An obvious choice of the cost function is the Euclidean distance between the centers of the bounding boxes. However, the function does not consider the size of the bounding box which can provide useful information for matching. Therefore, we project the locations and sizes of $\pazocal{P}$ in image $\pazocal{I}_{t+1}$ and obtain a set of estimated bounding boxes $\pazocal{\hat{P}}$. The cost function is then defined as the Jaccard index between the $\hat{p}$ and $q$:
 \begin{equation}
c(p, q) = \frac{\vert \hat{p} \cap q \vert}{\vert \hat{p} \cup q \vert}
\end{equation}
where $\hat{p}$ is the projection of $p$ in the next frame, and $\hat{p}$ is estimated using three techniques: 1) Lucas-Kanade optical flow, 2) Kalman filter, and 3) kernelized correlation filter.
 
Firstly, Lucas-Kanade optical flow is used to extrapolate the centers of $\pazocal{P}$. The algorithm calculates the motion $\mathbf{d}_{i,t} = \left[ d_{i,t}^x, \,d_{i,t}^y \right]^T$ for a given point $\mathbf{o}_{i,t} = \left[ o_{i,t}^x, \, o_{i,t}^y \right]^T$ between image pair $\left( \pazocal{I}, \, \pazocal{J} \right)$. The projected center for bounding box $p$, $\hat{\mathbf{o}}_{p,t+1}$, is then estimated with the mean of the optical flows of the $m$ feature points within the box: 

\begin{equation}
\hat{\mathbf{o}}_{p,t+1} = \mathbf{o}_{p,t} + \frac{1}{m}\sum_{i=1}^{m} \mathbf{d}_{i,t}
\end{equation}

The feature points can be extracted using any point feature algorithms such as SIFT, SURF, and Harris corner. To ensure the features focus on the object rather than background pixels in the bounding box, we extract and track features from the excess green image $\pazocal{G} = \frac{2G - R - B}{R + G + B}$ where R, G, and B are red, green, and blue channels of the original RGB image. However, features may become unavailable when objects are temporarily occluded. In addition, the robot usually moves at a steady speed, and thus using a dynamic model can improve the accuracy of feature tracking. Therefore, we robustify the optical flow with the Kalman filter. 

For bounding box $p$, we define the state vector at frame $t$ as $\mathbf{x}_{p,t} = \left[ o_{p,t}^x, \, \, o_{p,t}^y, \, \, \dot{o}_{p,t}^x, \, \, \dot{o}_{p,t}^y \right]^T$. The robot usually moves in constant velocity, so we consider a linear system model and define the state transition model and observation model as in Equation \eqref{eqn:system}:
\begin{equation}
\label{eqn:system}
\begin{aligned}
\mathbf{x}_{t+1} &=
\begingroup
\renewcommand*{\arraystretch}{0.5}
\begin{bmatrix}
1 & 0 & \Delta t & 0 \\
0 & 1 & 0 & \Delta t \\
0 & 0 & 1 & 0 \\
0 & 0 & 0 & 1
\end{bmatrix}
\endgroup
\mathbf{x}_t + \mathbf{\omega}_t \\
\mathbf{z}_{t+1} &= 
\begingroup
\renewcommand*{\arraystretch}{0.5}
\begin{bmatrix}
1 & 0 & 0 & 0 \\
0 & 1 & 0 & 0 \\
0 & 0 & 1 & 0 \\
0 & 0 & 0 & 1
\end{bmatrix}
\endgroup
\mathbf{x}_{t+1} + \mathbf{\nu}_t
\end{aligned}
\end{equation}
where $\Delta t$ is the time interval between $t$ a $t+1$.  $\mathbf{\omega}_t$ and $\mathbf{\nu}_t$ are process and measurement noise, both assumed to be random variables drawn from zero-mean Gaussian distribution as $\mathbf{\omega}\sim\pazocal{N}(0,\; \mathbf{Q})$ and $\mathbf{\nu}\sim\pazocal{N}(0,\; \mathbf{R})$, where $\mathbf{Q}$ and $\mathbf{R}$ are user-defined covariance matrices. The noisy measurement is taken from the optical flow estimates as:
\begin{equation}
\mathbf{z}_{p,t+1} = 
\begin{bmatrix}
\hat{\mathbf{o}}_{p,t+1} \\
\frac{1}{m}\sum_{i=1}^{m} \mathbf{d}_{i,t}
\end{bmatrix}
\end{equation}

After applying the standard Kalman filter prediction and update, we obtain the \textit{posteriori} state estimate $\tilde{\mathbf{x}}_{t+1}$ for each bounding box. The new center estimate is then taken as $\tilde{\mathbf{o}}_{p, t+1} = \left[ \tilde{o}^x_{p,t+1}\; \tilde{o}^y_{p,t+1} \right]^T $.

Finally, we apply a kernelized correlation filter (KCF)  \cite{henriques2015high} to refine the object location and account for scale changes due to perspective shift.

\subsection{Counting by detection and tracking}
The counting logic that combines the detection and tracking module is given in Algorithm~\ref{algo:counting}. At every time step \textit{t}, the detection network outputs a set of detection boxes. Meanwhile, all the trackers from the previous time step $t-1$ will update their projected location at \textit{t}. The algorithm tries to match each detection box with an existing tracker to resume the identity by maximizing the total Jaccard Index for all matches. Increment the count number if a tracker has found a match for more than a threshold frame and mark it as counted to avoid double counting. A tracker is also allowed to stay unmatched for a few frames before it gets destroyed. We found that tolerance provides protection against false detections and occlusions.

\IncMargin{0.7em}

\SetKwFunction{FRecurs}{FnRecursive}

\SetKwProg{Fn}{Function}{}{end}

\SetKwProg{FnReturn}{Function}{}{}

\SetKwComment{Comment}{}{}

\SetKwInOut{Input}{Input}

\SetKwInOut{Output}{Output}

\SetKwInOut{Parameter}{Parameter}

\SetKwInOut{Init}{Initialize}
\begin{algorithm}


\caption{Corn stand count by detection and tracking} \label{algo:counting}
\Input{$I_0, \ldots, I_n :=$ image frames \\}
\Output{$C :=$ corn stand count}
 \Parameter{$S_{min} :=$ Minimum Jaccard index threshold \\
		$p :=$ Tracking success threshold \\
		$q :=$ Tracker removal threshold \\
		$T :=$ Tracker list
	}

\ForEach{$I_i$}{
		$\left\lbrace box^D \right\rbrace  \leftarrow$ detection on $I_i$ \\
		$\left\lbrace box^T \right\rbrace \leftarrow$ tracking on $I_i$ \\
		Match $\left\lbrace box^D \right\rbrace$ and $\left\lbrace box^T \right\rbrace$ to maximize the total Jaccard index \\
		\ForEach{matched box pair $\left( box_m^D, box_m^T \right)$ } {
			$S \leftarrow$ Jaccard index for $\left( box_m^D, box_m^T \right)$\\
			\If{$S < S_{min}$} {
				remove $\left( box_m^D, box_m^T \right)$ from matched list
			}
			\Else{
				tracker $\leftarrow$ T[$box_m^T$] \\
				\If{tracker.counted is True}{
					continue
				}
				\Else{
					tracker.lost = 0 \\
					tracker.success++ \\
					\If{tracker.success $>$ p}{
						C++ \\
						tracker.counted = True
					}
				}
			}
		}
	}

	\ForEach{unmatched $box_u^T$} {
		tracker $\leftarrow$ T[$box_u^T$]\\
	tracker.lost++  \\
	 \If{tracker.lost $>$ q}{
			remove tracker from T
		}
	}

\ForEach{unmatched $box_u^D$} {	Initialize a tracker from $box_u^D$ and append to T	}

 \end{algorithm}

\section{Field Results} \label{sec_exp_res}

Field results of a low-cost, ultra-compact 3D printed and autonomous field robot for navigation algorithm, as well as deep learning-based corn stand counting algorithm, are presented in this section.

\subsection{Real-time Results for Navigation} \label{sec_exp_res_nav}

\begin{figure*}
	\subfigure[]{
		\includegraphics[width=0.33\linewidth]{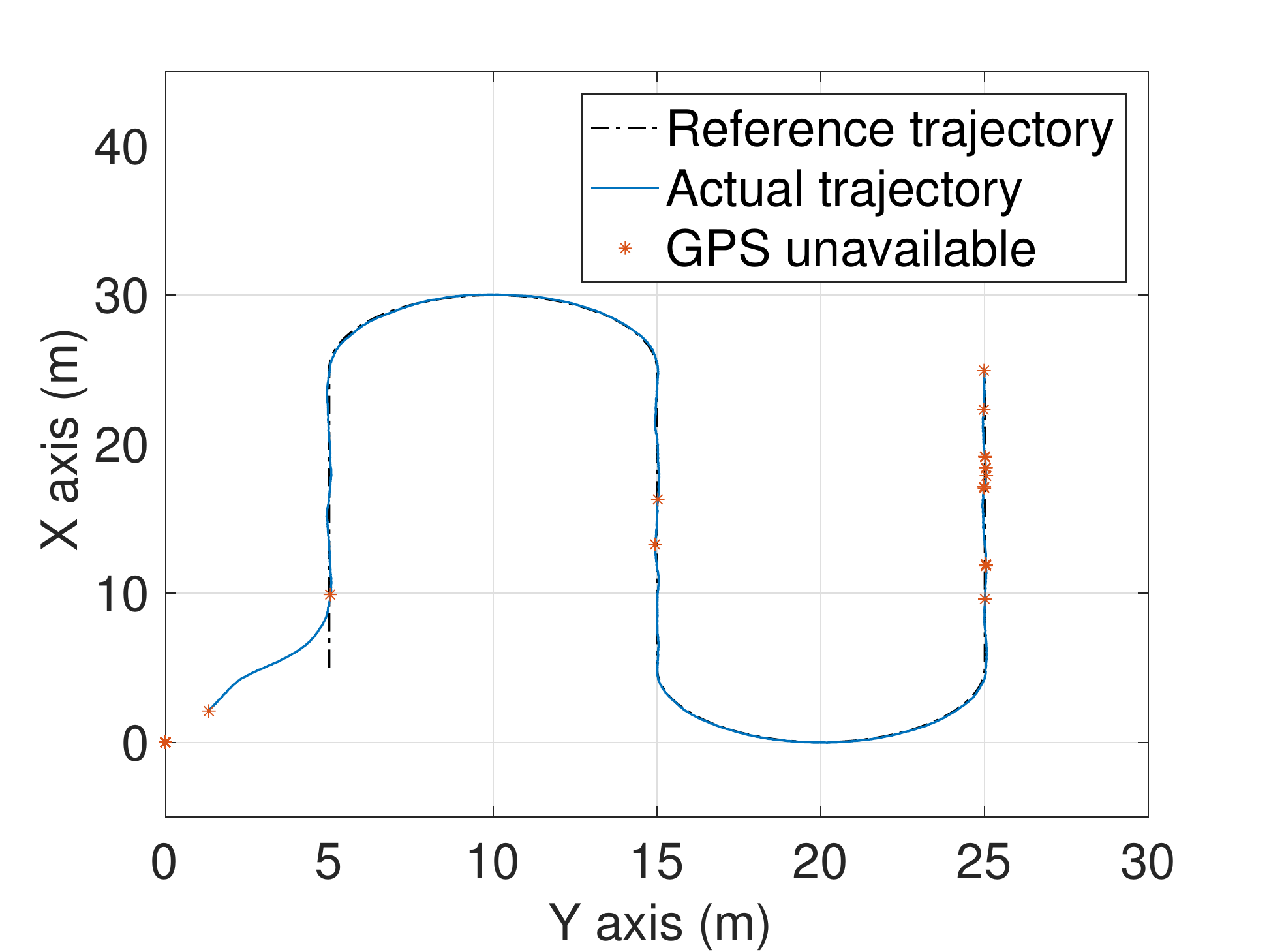}
\label{fig_tra}}	
	\subfigure[]{
		\includegraphics[width=0.33\linewidth]{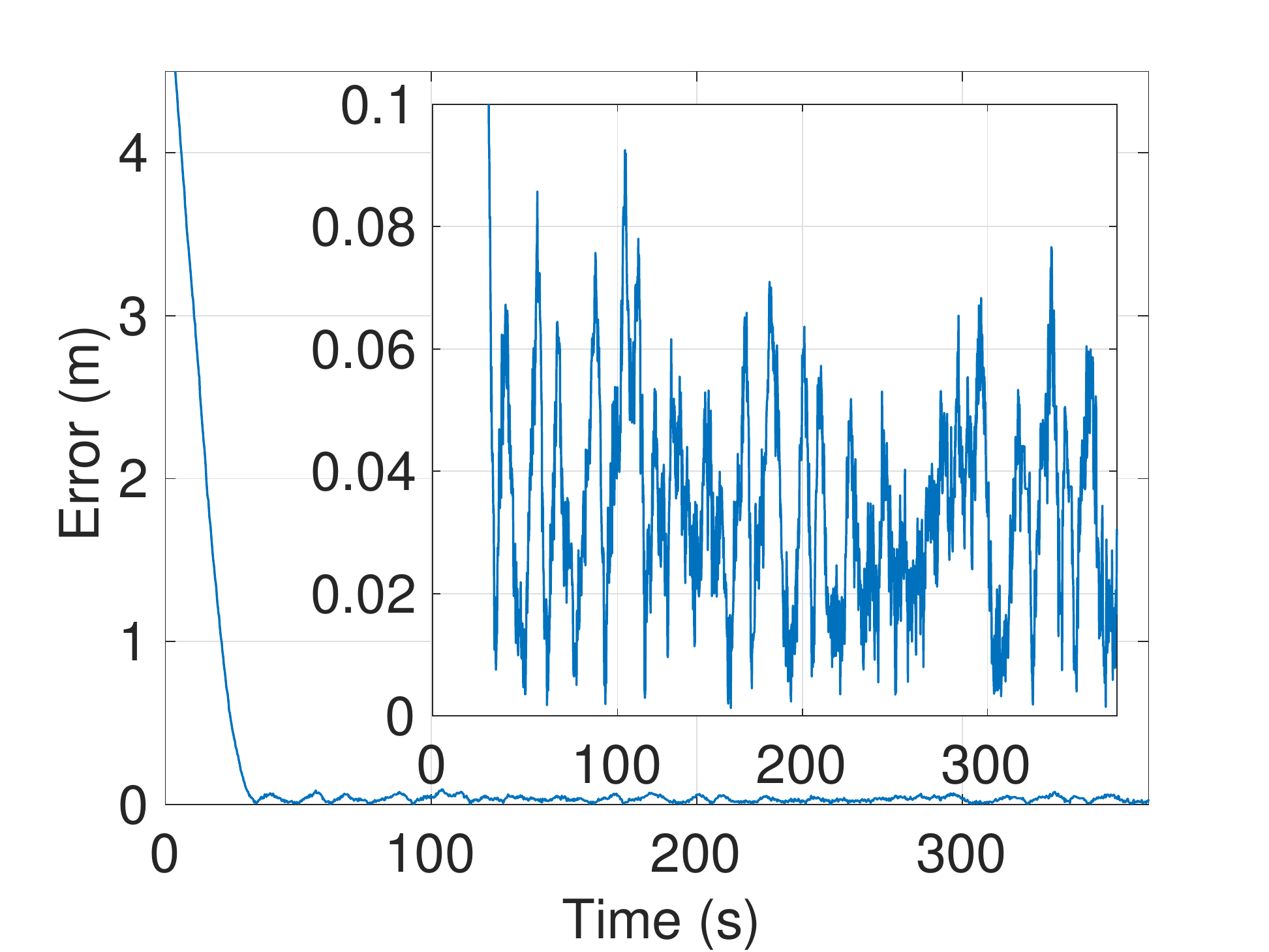}
	\label{fig_error}}
	\subfigure[]{
		\includegraphics[width=0.33\linewidth]{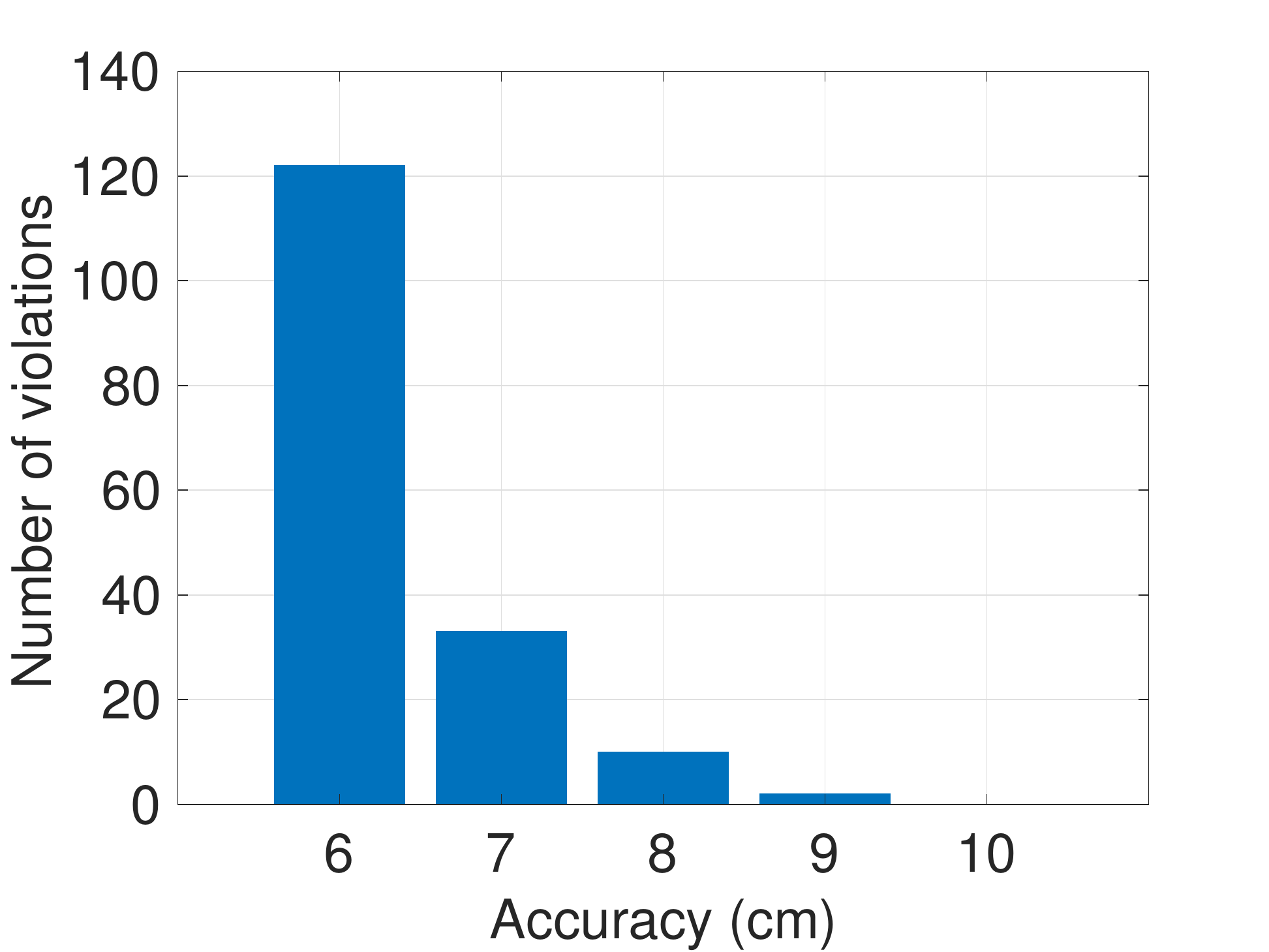}
	\label{fig_vio}}
		\subfigure[]{
		\includegraphics[width=0.33\linewidth]{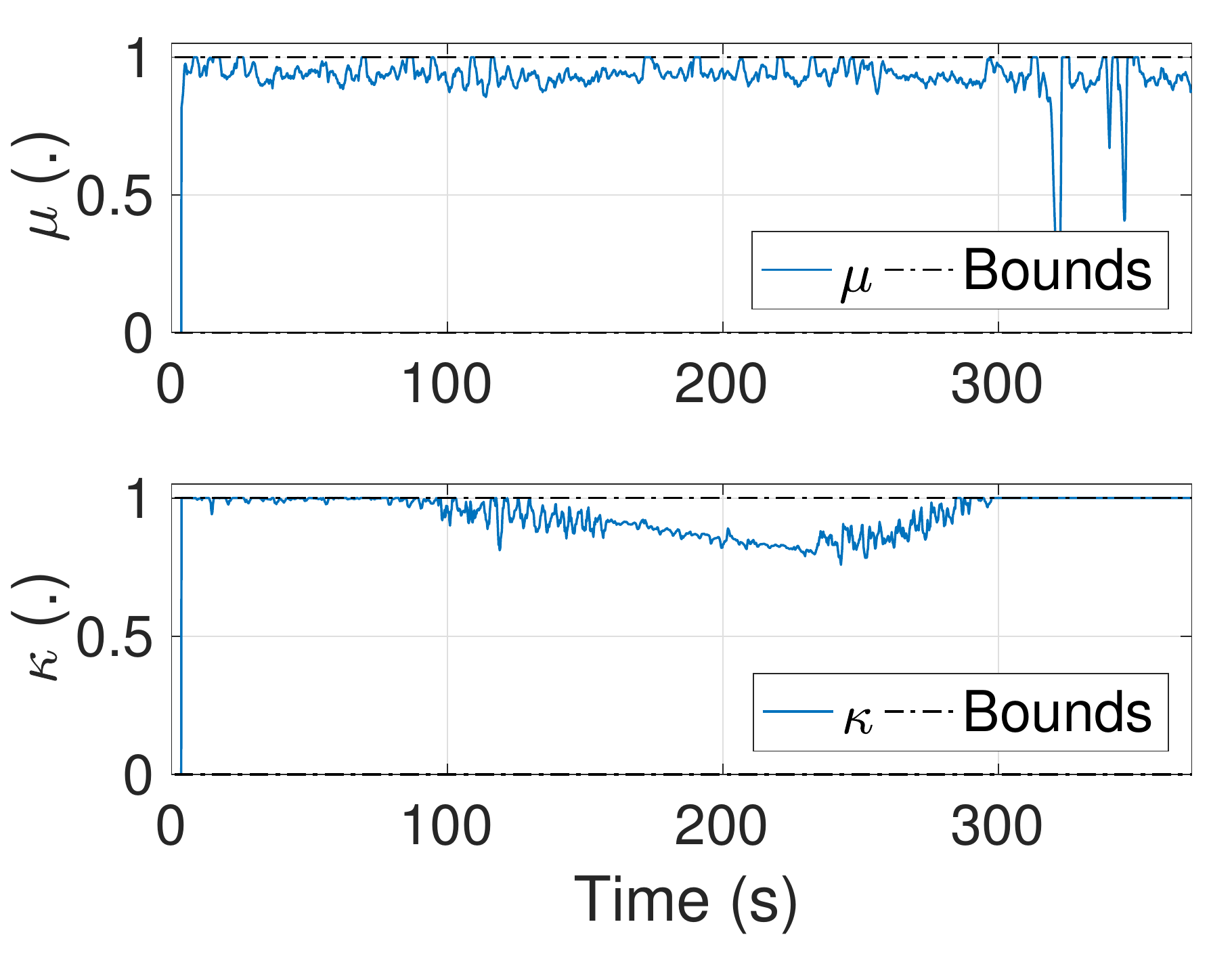}
	\label{fig_est}}
			\subfigure[]{
		\includegraphics[width=0.33\linewidth]{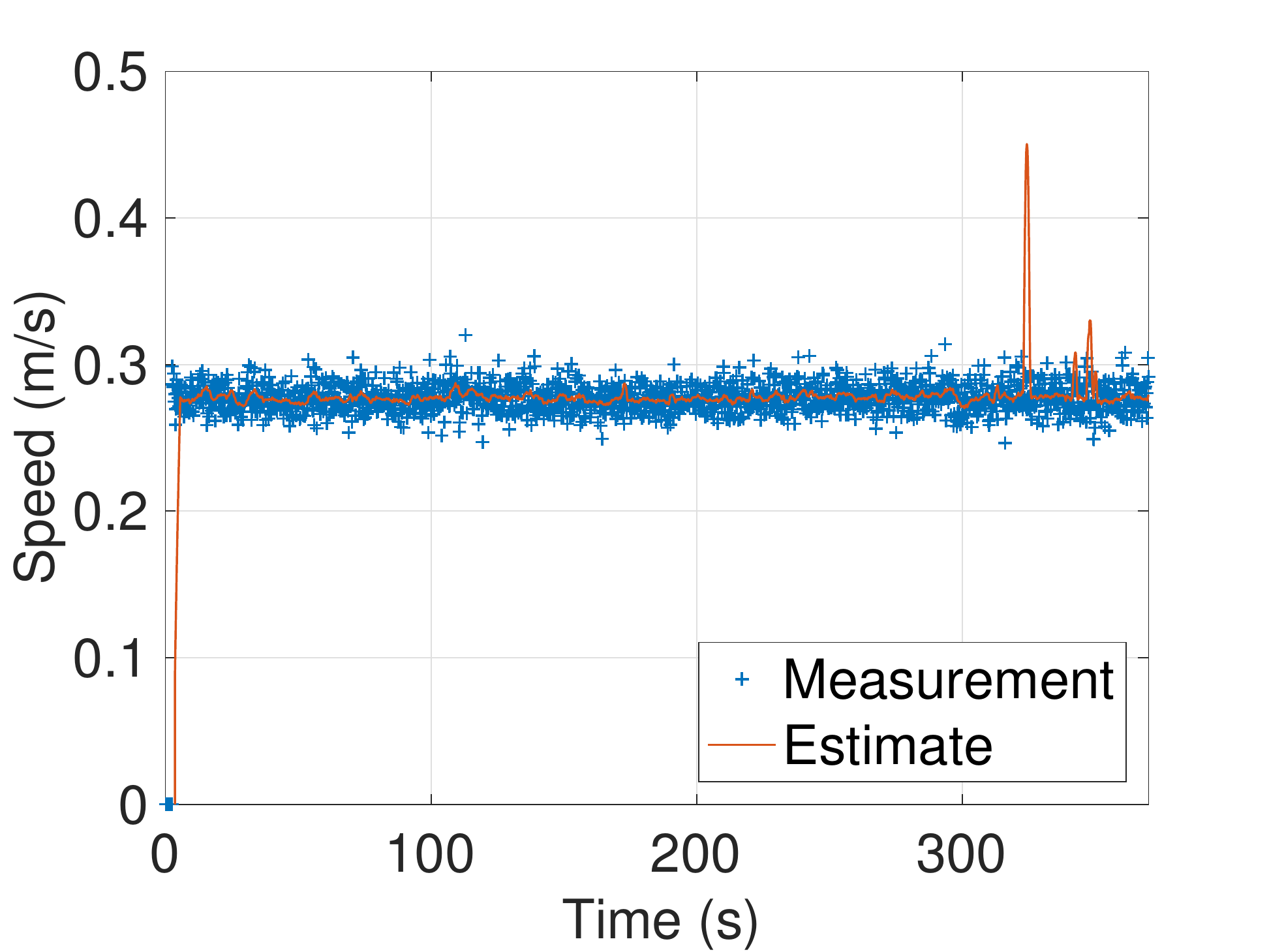}
		\label{fig_speed}}
\subfigure[]{
		\includegraphics[width=0.33\linewidth]{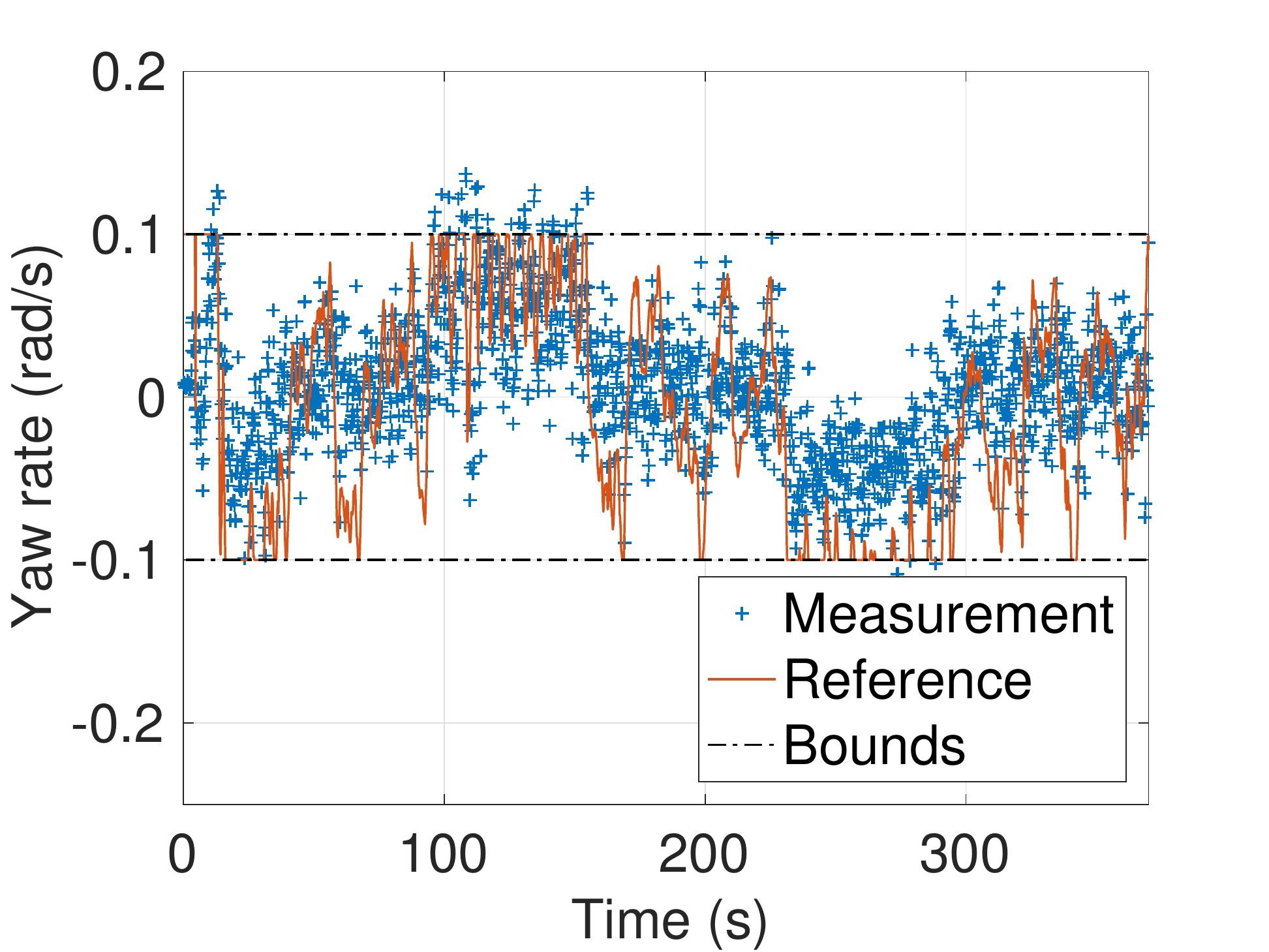}
	\label{fig_yawrate}}
	\caption{ a) Reference and actual trajectories, and unavailable Global Navigation Satellite System (GNSS) signals on the trajectory. b) Euclidean error. The mean value of Euclidean error is less than $5$ cm after being stayed on-track. c) The number of violations. The available space on either side of the field robot is restricted to 0.12 m. It does not violate this accuracy limit. d) Estimated traction parameters $\mu, \kappa$. e) Measured and estimated speeds. f) Reference and measured yaw rates. }
\end{figure*}
Linear controllers cannot achieve good trajectory tracking performance on uneven terrain if the system starts off-track \cite{Kayacan2018telc}. As shown in Fig. \ref{fig_tra}, after the 3D printed field robot is started off-track, it is capable of reaching to the reference trajectory and staying on-track. One of the major problems of such GNSS-based navigated outdoor robots is the missing data values from the satellites. This drawback is inevitable for Real-Time Kinematic (RTK) GNSSs, and the control algorithm has to be robust to cope with missing data points. As can be observed from Fig. \ref{fig_tra}, although there were 35 missing data points within 1850 data points throughout the experiment, high accurate trajectory tracking performance has been achieved by the NMHE-NMPC framework. 

The Euclidean error is shown in Fig. \ref{fig_error}. The mean value of the Euclidean error after staying on-track is 0.0423 m. The NMPC benefits from traction parameters estimated by the NMHE and results in highly accurate tracking performance so that the Euclidean error is less than $10$ cm. The number of violations is shown in Fig. \ref{fig_vio}. The results of multiple experiments indicate that the NMHE-NMPC framework does not violate this error constraint. This demonstrates the capability of the NMHE-NMPC framework.

The traction parameters estimates are shown in \ref{fig_est} and within the upper and lower bounds specified in \eqref{eq_nmpc}. It is clearly seen that they consistently stay close the upper bound and stabilize at values which ensure stable trajectory. It is to be noted that if the soil conditions are unchanging, then hard-coding might result in more accurate trajectory tracking performance. However, the soil conditions are dynamically changing in practice. 

The measured and estimated speeds of the ultra-compact 3D printed field robot are shown in Fig. \ref{fig_speed}. As can be seen, the robot has a constant speed, and the NMHE is capable of dealing with noise on the measurements. It is to be noted that some estimated speed values between 300 and 350 seconds are larger than measurements. The reason is that there are missing GNSS measurements as shown in Fig. \ref{fig_tra}. If GNSS measurement is invalid, previous valid measurement is fed to the NMHE instead of an outlier. There are 14 invalid points in the GNSS measurements between 318.8 and 321.6 seconds and these result in the peak values in the speed estimates between 322.8 and 325.6 seconds. The effect on the speed estimates arises seen subsequently because the speed in the system model \eqref{eq_systemmodel} is considered as a parameter and the parameter estimation is achieved in the arrival cost of the NMHE while the parameters in the estimation horizon are assumed to be time-invariant. Moreover, the longitudinal traction parameter estimate in Fig. \ref{fig_est} has nadir values between 322.8 and 325.6 seconds similar to the speed estimate. Consequently, the effective speed $\mu v$ is equal to the speed of the robot despite unavailable GNSS points. The same explanation is the case for the two peak points the speed estimate between after around 340 seconds. 

The output of the NMPC, i.e., the desired yaw rate, and yaw rate measurements are shown in Fig. \ref{fig_yawrate}. The generated control signal by the NMPC is within the lower and upper bounds specified in \eqref{eq_nmpc}. The performance of the low-level controller is sufficient to track the reference signals despite the high noisy measurements. 

\subsection{Real-Time Results for Corn Stand Counting }

We collected data on the cornfield in Champaign, Illinois on May 15 and 16, 2018. The robot was driven while recording data for 56 plots. The corn plants were between V2 and V3 according to the growth stage by Abendroth et al. \cite{abendroth2011corn}. We extracted 211 images from the videos of the first 3 plots to train (169 samples) and validate (42 samples) the Faster R-CNN model and used the rest 53 plots to test the performance of the algorithm.

\subsubsection{Object Detection}

We trained the Faster R-CNN model using the Tensorflow object detection API \cite{huang2017speed}. The base network was ResNet101 \cite{he2016deep} with parameters pre-trained on the COCO dataset \cite{lin2014microsoft}. The model was then fine-tuned on the training set with Gradient Descent optimizer for 100 epochs. The training samples were augmented by random perturbations including scaling, horizontal flipping, translating, etc. Despite the small sample size, the model showed high accuracy on the validation set. The mean average precision (mAP) evaluated at $0.5$ and $0.7$ intersection over union (IOU) was $0.998$ and $0.764$, respectively.

\subsubsection{Validation: In-Field Corn counting}

\begin{figure}
	\centering
  \includegraphics[width=\linewidth]{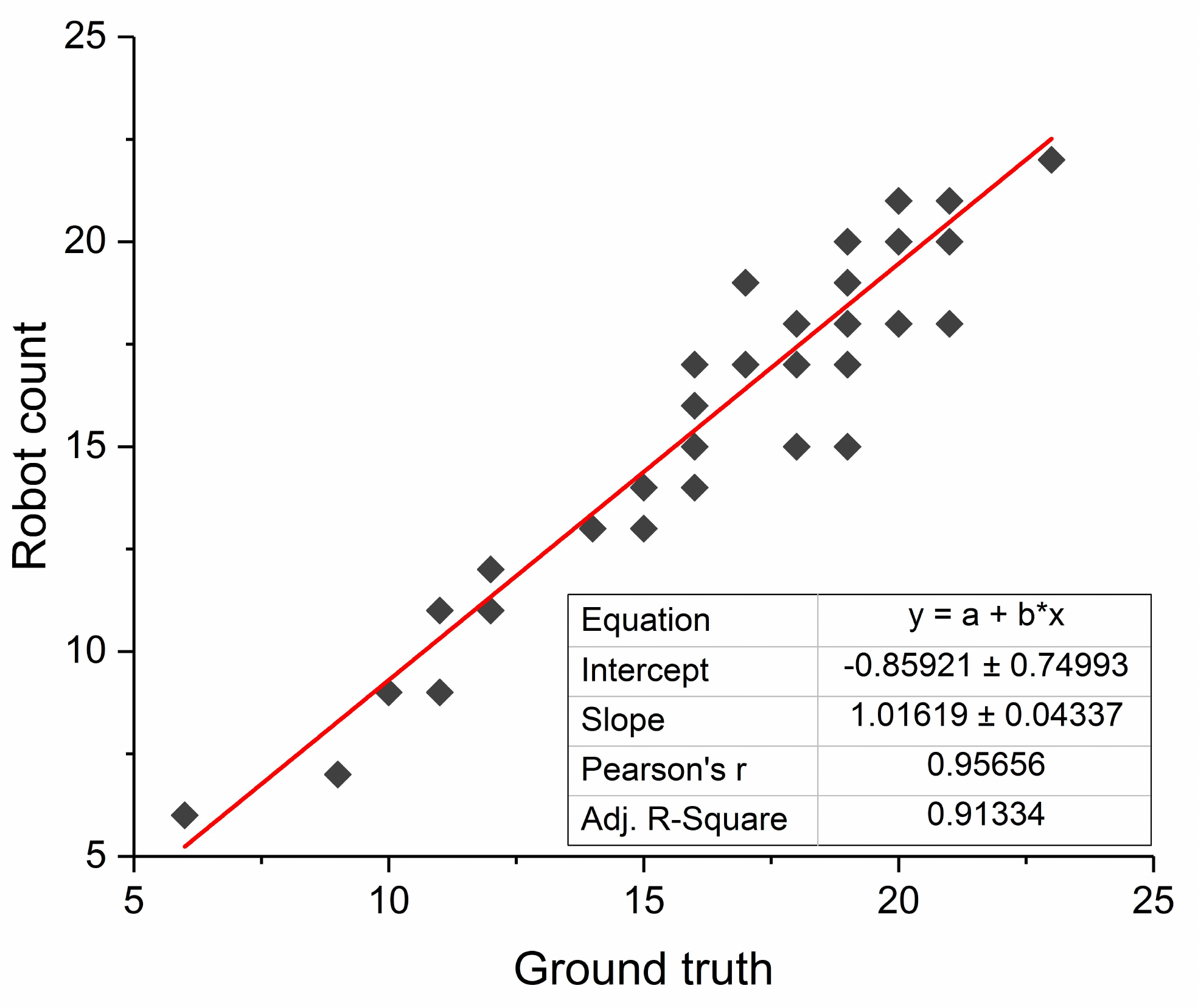}
\caption{ Scatter plot of corn plants per plot counted by robot vs human for.  
    The line represents a linear least-square fit. The robot predictions agree well with the ground truth with a correlation coefficient of $R=0.96$. }
 \label{fig:scatter plot} 
 \end{figure}

The accuracy of the whole counting algorithm was tested on the 53 plots that were not used for training the detection network. The corn plant population per plot obtained by robot vs human is compared in Fig. \ref{fig:scatter plot}. In general, the robot predictions agree well with the ground truth. The linear least-square fitted line through all data is given as $C_{robot}=1.02\times C_{human}-0.86$ and a Pearson's correlation coefficient $R=0.96$. The slope and intercept indicate that the algorithm approximates the identity function with systematic negative bias.

\begin{figure}
	\centering
  \includegraphics[width=\linewidth]{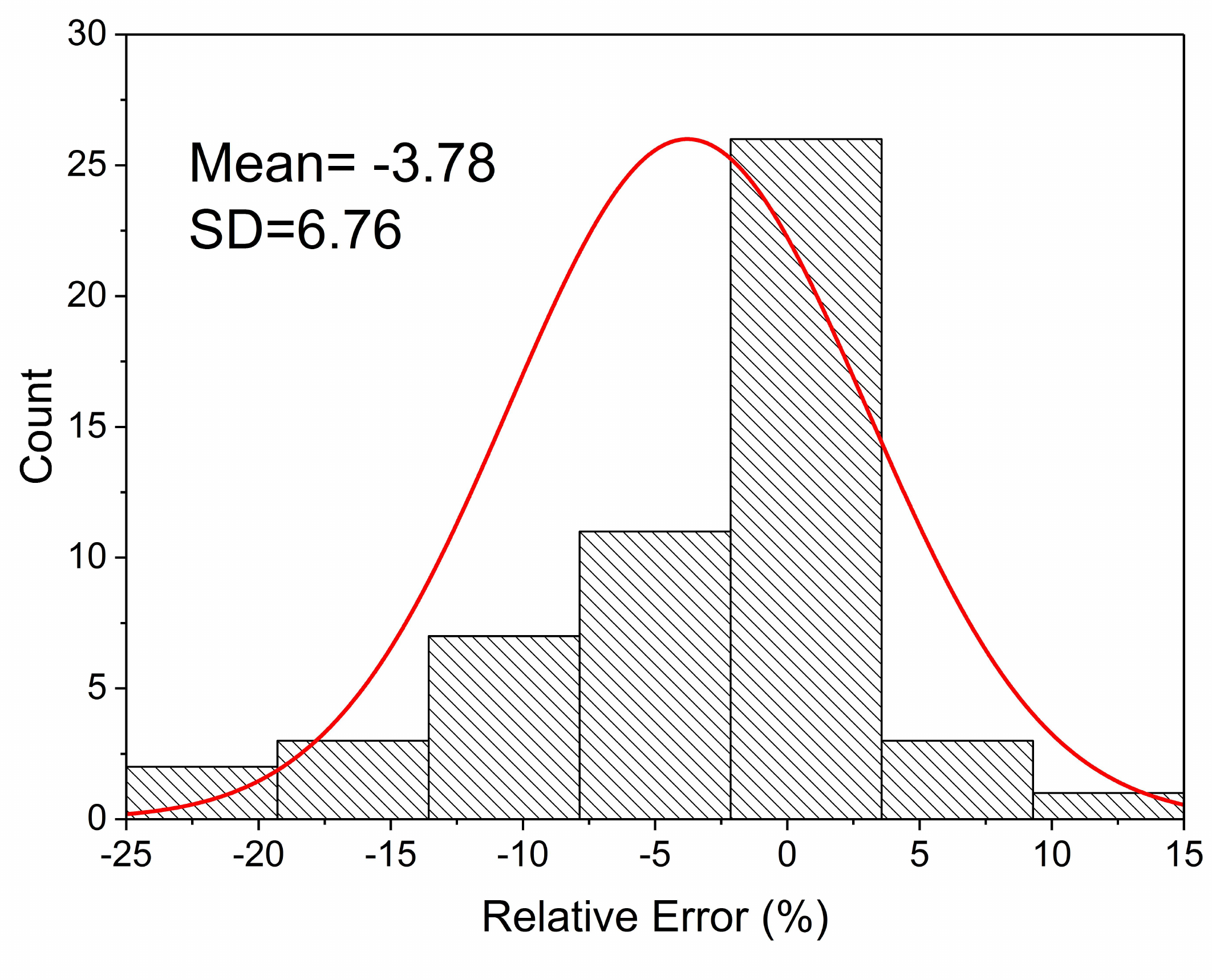}
 \caption{Histogram of counting errors. The red line shows a fitted Gaussian distribution with $\mu = -3.78$ and $\sigma = 6.76$.
 }
  \label{fig:histogram}
 \end{figure}

Figure \ref{fig:histogram} plots the histogram of the relative error $\epsilon$ defined in Equation \eqref{eqn:relative error}, where $C$ is the plant population. The mean of error is $-3.78\%$ and the standard deviation is $6.76\%$. The negative mean is consistent with the intercept of the linear fit, indicating that on average the algorithm tends to underestimate the population. After inspecting the output videos, we found that the errors largely attribute to tracking failure due to feature selection. Occasionally, features are unavailable to extract or extracted incorrectly from the background rather than the plants. In either case, the optical flow algorithm fails to predict the location in the next frame. Although Kalman filter mitigates the problem to some extent, small lags may occur between the detection and tracking results that can be accumulated and eventually lead to tracking failure. Currently, there is no mechanism for a tracker to recover from such failure. A potential solution is to allow the unmatched tracker to reconnect with a nearby detection within a distance based on the duration the tracker has remained matched. 

 \begin{equation}
\epsilon=\frac{C_{robot} - C_{human}}{C_{human}} \times 100\%
\label{eqn:relative error}
\end{equation}

\section{Conclusions}\label{sec_conc}

The real-time high precision control and deep learning-based corn stand counting algorithms for a low-cost, ultra-compact 3D printed, autonomous field robot for off-road terrain have been elaborated in this paper. Also, it specifically outlined the mechanical and wheel designs as well as the hardware and software architecture of the ultra-compact 3D printed field robot. 

An NMHE that identifies key terrain parameters using onboard robot sensors and a learning-based NMPC that enables highly accurate tracking have been developed to be capable of autonomously and safely navigating through row-based crops. The developed framework for high precision control that enables reliable field robot path tracking in off-road terrain provides less than $5$ cm accuracy in the presence of slip and does not violate the $12$ cm accuracy limit to avoid crop damage.


Additionally, we have presented an algorithm to estimate the corn plant population based on RGB imagery. The algorithm uses deep learning architecture (Faster R-CNN) to detect corn plants from images collected in field conditions.  Despite only using 169 training images, the detection model achieved mean Average Precision (mAP) of 0.998 and 0.764 at 0.5 and 0.7 Intersection over Union (IOU), respectively. The detected corn plants are then tracked using a novel visual tracking method that combines optical flow, Kalman filter, and kernelized correlation filter. In most cases, the tracker successfully maintains the identity of detected corn plants regardless of difficulties like perspective shift and motion blur. We have tested the accuracy of the algorithm on 53 plots with an average of 17 plants per plot. The robot predictions have agreed well with the ground truth with $C_{robot}=1.02\times C_{human}-0.86$ and a Pearson's correlation coefficient $R=0.96$. The mean relative error given by the algorithm is $-3.78\%$, and the standard deviation is $6.76\%$. The negative intercept and mean relative error indicate that the algorithm systematically underestimates the population primarily due to tracking failure. A future direction would be to include a mechanism that allows trackers to recover from failures based on their tracking history.

These results demonstrate a significant step towards autonomous field robot-based real-time phenotyping using low-cost, ultra-compact ground robots for corn plants and potentially other crops.

\section*{Acknowledgments}
The information, data, or work presented herein was funded in part by the Advanced Research Projects Agency-Energy (ARPA-E), U.S. Department of Energy, under Award Number DE-AR0000598. The views and opinions of authors expressed herein do not necessarily state or reflect those of the United States Government or any agency thereof.

Erkan Kayacan and Zhongzhong Zhang made equal contributions to this work.

Dr. Kayacan was a postdoctoral researcher in Chowdhary's group when the bulk of this work was undertaken. He is a Lecturer at the University of Queensland since April 2019. 

Any questions or comments should be directed to Girish Chowdhary. We thank UIUC-IGB TERRA-MEPP team and EarthSense Inc. for valuable suggestions.

\bibliographystyle{spbasic}      

\bibliography{ref}

\begin{thebibliography}{43}
\providecommand{\natexlab}[1]{#1}
\providecommand{\url}[1]{{#1}}
\providecommand{\urlprefix}{URL }
\expandafter\ifx\csname urlstyle\endcsname\relax
  \providecommand{\doi}[1]{DOI~\discretionary{}{}{}#1}\else
  \providecommand{\doi}{DOI~\discretionary{}{}{}\begingroup
  \urlstyle{rm}\Url}\fi
\providecommand{\eprint}[2][]{\url{#2}}

\bibitem[{Abendroth et~al.(2011)Abendroth, Elmore, Boyer, and
  Marlay}]{abendroth2011corn}
Abendroth LJ, Elmore RW, Boyer MJ, Marlay SK (2011) Corn growth and development

\bibitem[{Araus and Cairns(2014)}]{araus2014field}
Araus JL, Cairns JE (2014) Field high-throughput phenotyping: the new crop
  breeding frontier. Trends in plant science 19(1):52--61

\bibitem[{Biber et~al.(2012)Biber, Weiss, Dorna, and Albert}]{Biber2012}
Biber P, Weiss U, Dorna M, Albert A (2012) Navigation system of the autonomous
  agricultural robot bonirob. In: Workshop on Agricultural Robotics: Enabling
  Safe, Efficient, and Affordable Robots for Food Production (Collocated with
  IROS 2012), Vilamoura, Portugal

\bibitem[{Chen et~al.(2017)Chen, Shivakumar, Dcunha, Das, Okon, Qu, Taylor, and
  Kumar}]{chen2017counting}
Chen SW, Shivakumar SS, Dcunha S, Das J, Okon E, Qu C, Taylor CJ, Kumar V
  (2017) Counting apples and oranges with deep learning: A data-driven
  approach. IEEE Robotics and Automation Letters 2(2):781--788

\bibitem[{Das et~al.(2015)Das, Cross, Qu, Makineni, Tokekar, Mulgaonkar, and
  Kumar}]{das2015devices}
Das J, Cross G, Qu C, Makineni A, Tokekar P, Mulgaonkar Y, Kumar V (2015)
  Devices, systems, and methods for automated monitoring enabling precision
  agriculture. In: 2015 IEEE International Conference on Automation Science and
  Engineering (CASE), IEEE, pp 462--469

\bibitem[{Diehl et~al.(2002)Diehl, Bock, Schlöder, Findeisen, Nagy, and
  Allg{\"{o}}wer}]{Diehl2}
Diehl M, Bock H, Schlöder JP, Findeisen R, Nagy Z, Allg{\"{o}}wer F (2002)
  Real-time optimization and nonlinear model predictive control of processes
  governed by differential-algebraic equations. Journal of Process Control
  12(4):577 -- 585

\bibitem[{Ferreau et~al.(2012)Ferreau, Kraus, Vukov, Saeys, and
  Diehl}]{Ferreau}
Ferreau H, Kraus T, Vukov M, Saeys W, Diehl M (2012) High-speed moving horizon
  estimation based on automatic code generation. In: Decision and Control
  (CDC), 2012 IEEE 51st Annual Conference on, pp 687--692

\bibitem[{Furbank and Tester(2011)}]{furbank2011phenomics}
Furbank RT, Tester M (2011) Phenomics-technologies to relieve the phenotyping
  bottleneck. Trends in plant science 16(12):635--644

\bibitem[{Halstead et~al.(2018)Halstead, McCool, Denman, Perez, and
  Fookes}]{halstead2018fruit}
Halstead M, McCool C, Denman S, Perez T, Fookes C (2018) Fruit quantity and
  quality estimation using a robotic vision system. arXiv preprint
  arXiv:180105560

\bibitem[{Haseltine and Rawlings(2005)}]{Haseltine2005}
Haseltine EL, Rawlings JB (2005) Critical evaluation of extended kalman
  filtering and moving-horizon estimation. Industrial \& Engineering Chemistry
  Research 44(8):2451--2460, \doi{10.1021/ie034308l}

\bibitem[{He et~al.(2016)He, Zhang, Ren, and Sun}]{he2016deep}
He K, Zhang X, Ren S, Sun J (2016) Deep residual learning for image
  recognition. In: Proceedings of the IEEE conference on computer vision and
  pattern recognition, pp 770--778

\bibitem[{He et~al.(2017)He, Gkioxari, Doll{\'a}r, and Girshick}]{he2017mask}
He K, Gkioxari G, Doll{\'a}r P, Girshick R (2017) Mask r-cnn. In: Computer
  Vision (ICCV), 2017 IEEE International Conference on, IEEE, pp 2980--2988

\bibitem[{Henriques et~al.(2015)Henriques, Caseiro, Martins, and
  Batista}]{henriques2015high}
Henriques JF, Caseiro R, Martins P, Batista J (2015) High-speed tracking with
  kernelized correlation filters. IEEE Transactions on Pattern Analysis and
  Machine Intelligence 37(3):583--596

\bibitem[{van Henten et~al.(2009)van Henten, Goense, and Lokhorst}]{Henten2009}
van Henten EJ, Goense D, Lokhorst C (2009) Precision agriculture. Wageningen
  Academic Publishers, The Netherlands

\bibitem[{Huang et~al.(2017)Huang, Rathod, Sun, Zhu, Korattikara, Fathi,
  Fischer, Wojna, Song, Guadarrama et~al.}]{huang2017speed}
Huang J, Rathod V, Sun C, Zhu M, Korattikara A, Fathi A, Fischer I, Wojna Z,
  Song Y, Guadarrama S, et~al. (2017) Speed/accuracy trade-offs for modern
  convolutional object detectors. In: IEEE CVPR, vol~4

\bibitem[{Iagnemma et~al.(2004)Iagnemma, Kang, Shibly, and Dubowsky}]{1339393}
Iagnemma K, Kang S, Shibly H, Dubowsky S (2004) Online terrain parameter
  estimation for wheeled mobile robots with application to planetary rovers.
  IEEE Transactions on Robotics 20(5):921--927, \doi{10.1109/TRO.2004.829462}

\bibitem[{Kayacan and Chowdhary(2019)}]{Kayacan2018telc}
Kayacan E, Chowdhary G (2019) Tracking error learning control for precise
  mobile robot path tracking in outdoor environment. Journal of Intelligent
  {\&} Robotic Systems 95(3-4):975--986

\bibitem[{Kayacan et~al.(2013)Kayacan, Kayacan, Ramon, and Saeys}]{6606388}
Kayacan E, Kayacan E, Ramon H, Saeys W (2013) Modeling and identification of
  the yaw dynamics of an autonomous tractor. In: 2013 9th Asian Control
  Conference (ASCC), pp 1--6

\bibitem[{Kayacan et~al.(2016)Kayacan, Peschel, and Kayacan}]{erkan2016acc}
Kayacan E, Peschel JM, Kayacan E (2016) Centralized, decentralized and
  distributed nonlinear model predictive control of a tractor-trailer system: A
  comparative study. In: 2016 American Control Conference (ACC), pp 4403--4408,
  \doi{10.1109/ACC.2016.7525615}

\bibitem[{Kayacan et~al.(2018{\natexlab{a}})Kayacan, Kayacan, Chen, Ramon, and
  Saeys}]{Kayacan2018}
Kayacan E, Kayacan E, Chen IM, Ramon H, Saeys W (2018{\natexlab{a}}) On the
  Comparison of Model-Based and Model-Free Controllers in Guidance, Navigation
  and Control of Agricultural Vehicles, Springer International Publishing,
  Cham, pp 49--73

\bibitem[{Kayacan et~al.(2018{\natexlab{b}})Kayacan, Saeys, Ramon, Belta, and
  Peschel}]{8409989}
Kayacan E, Saeys W, Ramon H, Belta C, Peschel JM (2018{\natexlab{b}})
  Experimental validation of linear and nonlinear mpc on an articulated
  unmanned ground vehicle. IEEE/ASME Transactions on Mechatronics
  23(5):2023--2030

\bibitem[{Kayacan et~al.(2018{\natexlab{c}})Kayacan, Young, Peschel, and
  Chowdhary}]{KayacanJFR}
Kayacan E, Young SN, Peschel JM, Chowdhary G (2018{\natexlab{c}})
  High-precision control of tracked field robots in the presence of unknown
  traction coefficients. Journal of Field Robotics 35(7):1050--1062

\bibitem[{Kayacan et~al.(2018{\natexlab{d}})Kayacan, Zhang, and
  Chowdhary}]{KayacanRSS}
Kayacan E, Zhang Z, Chowdhary G (2018{\natexlab{d}}) Embedded high precision
  control and corn stand counting algorithms for an ultra-compact 3d printed
  field robot. In: Proceedings of Robotics: Science and Systems, Pittsburgh,
  Pennsylvania, \doi{10.15607/RSS.2018.XIV.036}

\bibitem[{Krizhevsky et~al.(2012)Krizhevsky, Sutskever, and
  Hinton}]{krizhevsky2012imagenet}
Krizhevsky A, Sutskever I, Hinton G (2012) Imagenet classification with deep
  convolutional neural networks. In: Advances in neural information processing
  systems, pp 1097--1105

\bibitem[{Lee and Iagnemma(2016)}]{7759528}
Lee SU, Iagnemma K (2016) Robust motion planning methodology for autonomous
  tracked vehicles in rough environment using online slip estimation. In: 2016
  IEEE/RSJ International Conference on Intelligent Robots and Systems (IROS),
  pp 3589--3594, \doi{10.1109/IROS.2016.7759528}

\bibitem[{Lee et~al.(2016)Lee, Gonzalez, and Iagnemma}]{7487413}
Lee SU, Gonzalez R, Iagnemma K (2016) Robust sampling-based motion planning for
  autonomous tracked vehicles in deformable high slip terrain. In: 2016 IEEE
  International Conference on Robotics and Automation (ICRA), pp 2569--2574,
  \doi{10.1109/ICRA.2016.7487413}

\bibitem[{Lin et~al.(2014)Lin, Maire, Belongie, Hays, Perona, Ramanan,
  Doll{\'a}r, and Zitnick}]{lin2014microsoft}
Lin TY, Maire M, Belongie S, Hays J, Perona P, Ramanan D, Doll{\'a}r P, Zitnick
  CL (2014) Microsoft coco: Common objects in context. In: European conference
  on computer vision, Springer, pp 740--755

\bibitem[{Liu et~al.(2015)Liu, Anguelov, Erhan, Szegedy, Reed, Fu, and
  Berg}]{Liu2015}
Liu W, Anguelov D, Erhan D, Szegedy C, Reed S, Fu CY, Berg AC (2015) Ssd:
  Single shot multibox detector. arXiv preprint arXiv:151202325

\bibitem[{Liu et~al.(2018)Liu, Chen, Aditya, Sivakumar, Dcunha, Qu, Taylor,
  Das, and Kumar}]{liu2018robust}
Liu X, Chen SW, Aditya S, Sivakumar N, Dcunha S, Qu C, Taylor CJ, Das J, Kumar
  V (2018) Robust fruit counting: Combining deep learning, tracking, and
  structure from motion. In: 2018 IEEE/RSJ International Conference on
  Intelligent Robots and Systems (IROS), IEEE, pp 1045--1052

\bibitem[{Liu et~al.(2019)Liu, Chen, Liu, Shivakumar, Das, Taylor, Underwood,
  and Kumar}]{liu2019monocular}
Liu X, Chen SW, Liu C, Shivakumar SS, Das J, Taylor CJ, Underwood J, Kumar V
  (2019) Monocular camera based fruit counting and mapping with semantic data
  association. IEEE Robotics and Automation Letters 4(3):2296--2303

\bibitem[{Long et~al.(2015)Long, Shelhamer, and Darrell}]{long2015fully}
Long J, Shelhamer E, Darrell T (2015) Fully convolutional networks for semantic
  segmentation. In: Proceedings of the IEEE Conference on Computer Vision and
  Pattern Recognition, pp 3431--3440

\bibitem[{Lu et~al.(2017)Lu, Cao, Xiao, Zhuang, and Shen}]{lu2017tasselnet}
Lu H, Cao Z, Xiao Y, Zhuang B, Shen C (2017) Tasselnet: counting maize tassels
  in the wild via local counts regression network. Plant methods 13(1):79

\bibitem[{Mayne et~al.(2000)Mayne, Rawlings, Rao, and Scokaert}]{Mayne}
Mayne D, Rawlings J, Rao C, Scokaert P (2000) Constrained model predictive
  control: Stability and optimality. Automatica 36(6):789 -- 814

\bibitem[{Mehndiratta et~al.(2019)Mehndiratta, Kayacan, Patel, Kayacan, and
  Chowdhary}]{Mehndiratta2019}
Mehndiratta M, Kayacan E, Patel S, Kayacan E, Chowdhary G (2019) Learning-Based
  Fast Nonlinear Model Predictive Control for Custom-Made 3D Printed Ground and
  Aerial Robots, Springer International Publishing, Cham, pp 581--605

\bibitem[{Mueller-Sim et~al.(2017)Mueller-Sim, Jenkins, Abel, and
  Kantor}]{7989418}
Mueller-Sim T, Jenkins M, Abel J, Kantor G (2017) The robotanist: A
  ground-based agricultural robot for high-throughput crop phenotyping. In:
  2017 IEEE International Conference on Robotics and Automation (ICRA), pp
  3634--3639, \doi{10.1109/ICRA.2017.7989418}

\bibitem[{Rahnemoonfar and Sheppard(2017)}]{rahnemoonfar2017deep}
Rahnemoonfar M, Sheppard C (2017) Deep count: fruit counting based on deep
  simulated learning. Sensors 17(4):905

\bibitem[{Ray(2009)}]{4840546}
Ray LE (2009) Estimation of terrain forces and parameters for rigid-wheeled
  vehicles. IEEE Transactions on Robotics 25(3):717--726,
  \doi{10.1109/TRO.2009.2018971}

\bibitem[{Ren et~al.(2015)Ren, He, Girshick, and Sun}]{Ren2015}
Ren S, He K, Girshick R, Sun J (2015) Faster r-cnn: Towards real-time object
  detection with region proposal networks. In: Advances in neural information
  processing systems, pp 91--99

\bibitem[{Shafiekhani et~al.(2017)Shafiekhani, Kadam, Fritschi, and
  DeSouza}]{Shafiekhani2017}
Shafiekhani A, Kadam S, Fritschi FB, DeSouza GN (2017) Vinobot and vinoculer:
  Two robotic platforms for high-throughput field phenotyping. Sensors 17(1)

\bibitem[{Simonyan and Zisserman(2014)}]{simonyan2014very}
Simonyan K, Zisserman A (2014) Very deep convolutional networks for large-scale
  image recognition. arXiv preprint arXiv:14091556

\bibitem[{Stein et~al.(2016)Stein, Bargoti, and Underwood}]{stein2016image}
Stein M, Bargoti S, Underwood J (2016) Image based mango fruit detection,
  localisation and yield estimation using multiple view geometry. Sensors
  16(11):1915

\bibitem[{Szegedy et~al.(2015)Szegedy, Liu, Jia, Sermanet, Reed, Anguelov,
  Erhan, Vanhoucke, and Rabinovich}]{szegedy2015going}
Szegedy C, Liu W, Jia Y, Sermanet P, Reed S, Anguelov D, Erhan D, Vanhoucke V,
  Rabinovich A (2015) Going deeper with convolutions. In: Proceedings of the
  IEEE conference on computer vision and pattern recognition, pp 1--9

\bibitem[{Young et~al.(2019)Young, Kayacan, and Peschel}]{Young2018}
Young SN, Kayacan E, Peschel JM (2019) Design and field evaluation of a ground
  robot for high-throughput phenotyping of energy sorghum. Precision
  Agriculture 20(4):697--722

\end{thebibliography}
\bibliographystyle{spbasic}

\end{document}